\documentclass[sigconf,screen]{acmart}

\AtBeginDocument{%
  }

\setcopyright{rightsretained}

\usepackage{graphicx, amsmath} 
\usepackage{latexsym}
\usepackage{lipsum}
\usepackage{array}
\usepackage{subcaption}
\usepackage{pifont}
\newcommand{\cmark}{\ding{51}}%
\newcommand{\xmark}{\ding{55}}%
\usepackage{booktabs}
\usepackage{listings}
\usepackage{bm}
\usepackage{fix-cm}
\usepackage{todonotes}
\usepackage{tabularx}
\usepackage{arydshln}
\usepackage{multirow}
\usepackage{makecell}
\usepackage{xspace}
\usepackage{cleveref} 
\usepackage{tablefootnote}

\begin{document}
\lstset{
  language=Python,
  basicstyle=\ttfamily\tiny,
  commentstyle=\color{gray},
  keywordstyle=\color{blue},
  stringstyle=\color{orange},
  numbers=left,
  numberstyle=\tiny,
  numbersep=5pt,
  breaklines=true,
  breakatwhitespace=true,
  tabsize=4,
  captionpos=b
}


\newcommand{\acr}{\textsl{COntext COMpression}\xspace}

\newcommand{\acrs}{\textsl{COCOM}\xspace}



\title{Context Embeddings for \\ Efficient Answer Generation in RAG}
\title{Context Embeddings for Efficient Answer Generation in RAG}

\author{David Rau}

\authornote{Equal Contribution.}
\affiliation{%
  \institution{University of Amsterdam}
  \city{Amsterdam}
   \country{Netherlands}
}
\email{d.m.rau@uva.nl}
\author{Shuai Wang}

\authornotemark[1]
\authornote{Work performed during an internship at Naver Labs Europe.}

\affiliation{%
  \institution{The University of Queensland}
  \city{Brisbane}
  \country{Australia}
}
\email{shuai.wang2@uq.edu.au}

\author{Herv\'{e} D\'{e}jean}

\affiliation{%
  \institution{Naver Labs Europe}
  \city{Grenoble}
  \country{France}
}
\email{herve.dejean@naverlabs.com}

\author{St\'{e}phane Clinchant}

\affiliation{%
  \institution{Naver Labs Europe}
  \city{Grenoble}
  \country{France}
}
\email{stephane.clinchant@naverlabs.com}










\renewcommand{\shortauthors}{Rau et al.}

\begin{abstract}
\textit{Retrieval-Augmented Generation (RAG)}  allows overcoming the limited knowledge of LLMs by extending the input with external information. As a consequence, the contextual inputs to the model become much longer which slows down decoding time directly translating to the time a user has to wait for an answer. We address this challenge by presenting \acrs, an effective context compression method, reducing long contexts to only a handful of \textit{Context Embeddings} speeding up the generation time by a large margin.
Our method allows for different compression rates trading off decoding time for answer quality. 
Compared to earlier methods, \acrs~ allows for handling multiple contexts more effectively, significantly reducing decoding time for long inputs.
Our method demonstrates an inference speed-up of up to 5.69$\times$ while achieving higher performance compared to existing efficient context compression methods.
Model checkpoints: \url{https://huggingface.co/naver/cocom-v1-128-mistral-7b}.

\end{abstract}

\maketitle

\section{Introduction}

\begin{figure}
  \includegraphics[width=\linewidth]{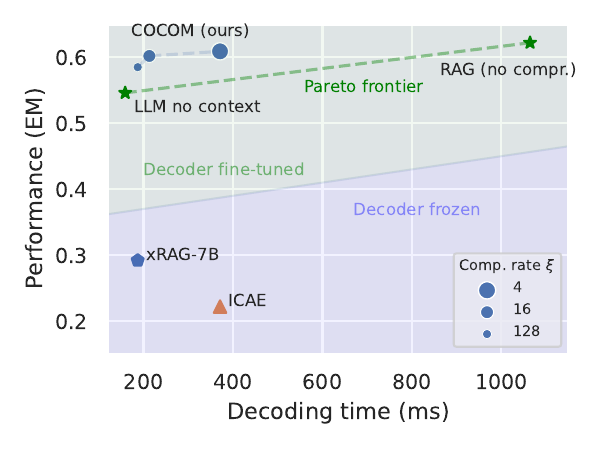}
     \caption{\acrs: Compressing multiple contexts for RAG into a small set ($\xi = {4, 16, 128}$) of \textit{Context Embeddings} leads to a massive speed up in answer generation while maintaining higher performance compared to other methods. Results are shown for the ASQA dataset.}
  \label{fig:teaser}
\end{figure}

Large Language Models (\emph{LLMs}) are pre-trained on massive amounts of textual data; for instance, Llama 2 \cite{touvron2023llama} has been trained on 3 trillion tokens during pre-training. Through billions of learnable parameters, LLMs not only excel at modeling language but at the same time, build up a knowledge base that could be later used for question answering. On the other hand, the model is limited to the knowledge contained in the pre-training data.
In knowledge-intensive scenarios, relying solely on the parametric memory of the model is often insufficient. To alleviate this, context can be provided \emph{explicitly} from an external source through a preceding retrieval step (Retrieval-Augmented Generation--\emph{RAG}). Although LLMs show notable improvements when given additional relevant context in knowledge-intensive tasks, this approach has limitations.

A key drawback is that adding more context to the input considerably slows down generation during inference. This occurs because the self-attention mechanism in transformers grows exponentially in space and memory requirements with increasing input length. 
At the same time, previous research has shown providing multiple documents as context can improve RAG performance \cite{izacard_atlas_2022, hsia2024ragged}. This is particularly critical for QA applications where reasoning over context from multiple documents is necessary, such as in multi-doc QA tasks \cite{fan-etal-2019-eli5, joshietal2017-triviaqa, yang-etal-2018-hotpotqa}.
In fact, the observation that modern transformers can naturally cope with many context documents for answer generation in open domain QA tasks was central to the development of RAG  \citep{DBLP:conf/wsdm/DehghaniAKR19,izacard_leveraging_2021}.
However, as the input length becomes larger, the position bias in LLMs might further complicate the extraction of relevant information \cite{liu_lost_2023}.

Previous work has shown that the increased generation time in RAG can be alleviated by reducing the model's input through context compression. This can be achieved either by applying lexical-based compression, where unimportant terms or tokens in the context are identified and filtered out during generation~\cite{jiang-etal-2023-llmlingua}, or by embedding-based compression, where embedding models transform the context into fewer embedding tokens in the LLM input~\cite{ge2024incontext,tan2024lloco,cheng2024xrag,muennighoff2024generative}. Notably, state-of-the-art embedding-based compression methods often achieve higher effectiveness and lower latency compared to lexical-based compression methods~\cite{cheng2024xrag}.

However, despite the current embedding-based compression approaches achieving lower latency in RAG systems, several limitations remain:
\begin{itemize}
    \item \textbf{Large compressor model}: These methods rely on large compression models to achieve high effectiveness, such as  \cite{cheng2024xrag, muennighoff2024generative}.

    \item \textbf{Low effectiveness}: The effectiveness of current embedding-based compression methods underestimates the potential of LLMs for answer generation, as they only tune parts of model components and leave the decoder LLM untuned. We hypothesize that freezing the decoder hinders the use of compressed contexts. 

    \item \textbf{Fixed compression rate}: Current methods do not offer different compression rates with respect to the length of input context, allowing to trade of inference time for generation quality at high effectiveness.
    \item \textbf{Single document limitation}: Current effective methods only support using a single document context to generate answers.
\end{itemize}

 We address the described limitations, similar to concurrently developed methods, by compressing contexts into a small number of \textit{context embeddings} which are then provided as input to the LLM.  This allows us to reduce the input size to a fraction of its surface form, which leads to an increased decoding time during answer generation. We call our model \textbf{\acrs~} (\textbf{CO}ntext \textbf{CO}mpression \textbf{M}odel), a multi-context compression method leveraging a single model for context compression and answer generation.
 

Additionally,
we further show that with appropriate pretraining and tuning approaches, our compressing model achieves significantly higher effectiveness than current context compressing approaches (see Figure \ref{fig:teaser}). We summarize our contributions as follows: 
    
\begin{itemize}
    \item We present \acrs, an effective context compression method, reducing long contexts to only a handful of context embeddings speeding up the generation time while achieving higher performance compared to other methods.
    \item In an efficiency study, we demonstrate the efficiency-effectiveness trade-offs achievable with different compression rates. We further illustrate the time and memory required for compression. We reduce inference time by up to 5.69 $\times$ and GFLOPs by up to 22 $\times$ while maintaining high performance.
    \item We conduct an ablation to understand which factors are the most important for effective generation and analyze the impact of the pretraining collection, pretraining, fine-tuning, and freezing or not the decoder. on the target dataset, and training the decoder.
\end{itemize}

The rest of this paper is structured in the following way.
\Cref{sec:related_work} discusses related work on RAG, efficiency, and compression approaches.
We continue in \Cref{sec:met} discussing the RAG task and our novel \acrs approach to effective context compression.
\Cref{sec:exp} details the experimental setup in terms of the RAG models and the five QA tasks.
In \Cref{sec:res}, we present the main \acrs results in terms of effectiveness and efficiency.
\Cref{sec:ana} conducts further analysis of how compression affects the model.
We end with discussion and conclusions in \Cref{sec:con}, and limitations in \Cref{sec:lim}.

\begin{table*}[!t]
     \caption{Comparison to previous works on Embedding-based Context Compression.}
    \label{tab:comparison_literature}
    \centering
    \setlength\tabcolsep{0pt} 
    \begin{tabular*}{\linewidth}{@{\extracolsep{\fill}} l *{5}{c}}
    \toprule
   \bf Work &  \bf Light Compressor & \bf Decoder Tuning & \bf Adaptable \bf $\gamma$ & \bf Multi-Doc & \bf Efficient Answer Generation \\
     \midrule
       GridLM \cite{muennighoff2024generative} & \xmark & \cmark & \xmark & \xmark & \xmark \\ 
          AutoCompressor \cite{chevalier2023adapting} & \xmark & \cmark & \cmark & \xmark & \cmark\\
    ICAE \cite{ge2024incontext} & \xmark & \xmark & \cmark & \cmark & \cmark \\

    xRAG \cite{cheng2024xrag} & \xmark & \xmark & \xmark & \xmark & \cmark \\
     \acrs-light~\textit{(ours)} & \cmark & \cmark & \cmark & \cmark & \cmark \\
         \acrs~\textit{(ours)} & \xmark & \cmark & \cmark & \cmark & \cmark \\
    \bottomrule
    \end{tabular*}
\end{table*}

\section{Related Work}
\label{sec:related_work}
In this section, we discuss related work on RAG, efficiency, and compression approaches.

The initial motivation for this work stems from a recent study by \citet{morris-etal-2023-text}, which demonstrates that a bag-of-words representation of the original surface terms can be recovered from text embeddings. 
This observation that embeddings can encapsulate the content of an entire passage inspired the idea to provide context in the form of an embedding rather than the original context in token form to an LLM. 

The underlying motivation in the context of RAG to reduce the input size is, as mentioned earlier, due to the computational costs of contextualizing long inputs and as a consequence thereof increased decoding time  \cite{asai2024reliable}. We address this by reducing the provided context to only a handful of context embeddings that are provided the LLM head-on. 

Reducing the input to RAG models is a very active research field, with many works being done concurrently with ours. Among those works, two primary lines of research have emerged:
\textit{embedding-based} and \textit{lexical-based} context compression. We discuss them in the following.

\subsection{Lexical-based Compression.} 
Lexical-based compression focuses on either selecting tokens from the context \cite{li2023unlocking} or summarizing contexts \cite{xu2023recomp}, both aiming to retain essential information while reducing overall context size. LLMLingua comprises a query-independent token filtering module that uses a LLM to first select important tokens in the context. Then, a query-dependent token classifier is used to select tokens to form the compressed context. 

On the other hand, \citet{zhu2024accelerating} do not consider compression at the term level, but at the document level. Retrieved documents are either included or excluded with respect to the query. Only the included documents form the context for answer generation.
It is worth noting that current lexical-based compression approaches all rely on specific query inputs. Therefore, compression needs to be (partially) processed online not allowing to compress documents offline, slowing down generation time.

\subsection{Embedding-based Compression.}
Embedding-based compression approaches focus on compressing the context into one or multiple summary embeddings that can be directly interpreted by the decoder model.
This first work of this line is called AutoCompressor~\cite{chevalier2023adapting}. This approach attempts to compress contextual information by segmenting it into randomly segmented chunks, subsequently aggregating these into summary embeddings through an iterative process until target embedding size is met. However, the training of the summary embeddings relies exclusively on next token prediction tasks, raising concerns about their ability to effectively encapsulate relevant contextual data. Furthermore, AutoCompressor is designed primarily for long contexts, generating a minimum of 50 summary embeddings. Such a configuration is not suitable for common RAG pipelines where short passages are retrieved, such as KILT.

Building up on AutoCompressor, ICAE by~\citet{ge2024incontext} explores training a context compressor using the same LLM as the decoder model, and compress only once to get the summary embeddings. 
However, their approach limits the model's capacity by using a frozen decoder module, preventing the accumulation of gradients from the decoder part during training. In this paper, we argue that decoder training is an important factor that strongly impacts the performance of the model. We illustrate this argument in Section~\ref{sec:exp_pretraining}.

Furthermore, GridLM \citet{muennighoff2024generative} addresses the issue of double decoding the same context first for retrieval and then again as the provided context to the LLM. They use the same LLM for ranking and generation which allows them to cache all representations during encoding the contexts and to reuse them during generation. This approach compared to ours is limited to only a single context, does not speed up decoding time, and results in gigantic storage requirements.

\citet{cheng2024xrag} propose xRAG concurrently to our method. They directly reuse frozen ranking representations based on embedding models while freezing the decoder. Although this approach successfully resolves the double decoding problem, it suffers from low effectiveness because the representation is not trained prior to its application to compression tasks. This issue becomes particularly challenging when light-weight encoder models, such as DPR with 109 million parameters, are used as compressors. In such cases, the model achieves similar effectiveness to the Mistral-7b model when retrieval is not applied~\footnote{By default, xRAG uses a 7B SFR LLM-based ranking model as compressor}.
On the other hand, using retrieval representations from lightweight models for compression is counter-intuitive. Representations gathered from retrieval tasks may lack sufficient information to fully recover the context. Conversely, representation learned for compression demonstrate its capacity to reconstruct the original context~\cite{ge2024incontext}. This suggest that, upon further adjustment, it may show a higher potential to serve as an effective retriever.

\subsection{Overview}
In Table \ref{tab:comparison_literature} we contrast our method with the described related works on embedding-based compression. It is important to note that most previous works mentioned so far have only considered cases that may not directly apply to RAG settings but rather to long-context question answering. In their setting, only \textit{one relevant} document is used for each query to fulfill the user request. 

Therefore, such models are not naturally able to deal with effectively multiple documents. Furthermore, their reported effectiveness may not directly indicate the final performance in RAG systems, where the document may be potentially irrelevant, and often multiple top-retrieved documents are used. As a decoder model, by design, should be able to handle multiple context representations, we argue \textit{that fine-tuning the decoder} is a simple yet necessary solution compared to existing works



\section{Methodology}
\label{sec:met}

In this section, we detail the RAG task and our novel \acrs approach to effective context compression.

\subsection{Task Definition: RAG}
\label{sec:task}
RAG employs a ranking system $\mathcal{R}$ and a parametric generative language model  $\theta_{LLM}$, where the ranking system can be multi-staged. 
First, the ranking system builds a search index $\mathcal{I}$ based on a collection. Then, at request time, the index $\mathcal{I}$ is searched yielding context segments\footnote{The segments can be at different granularities for instance sentences, passages, or entire documents. In this work, we focus on passages.} $\mathcal{C}$ that are relevant to the user input $x$: $f_{\mathcal{I}, \mathcal{R}} :  \{x\} \rightarrow  \mathcal{C}$.

Next, the LLM generates a response $r$ based on the context $\mathcal{C}$ and user input $x$:
\begin{equation}
     \theta_{LLM} : \{\mathcal{C},x\}  \rightarrow r
\end{equation}
Note how in RAG the context is added to the input of the LLM dramatically increasing the input to the LLM, as $|\mathcal{C}| \gg |x|$.

\subsection{\acrs: Effective Context Compression }

\begin{figure*}[t]
    \centering
    \includegraphics[width=.8\textwidth]{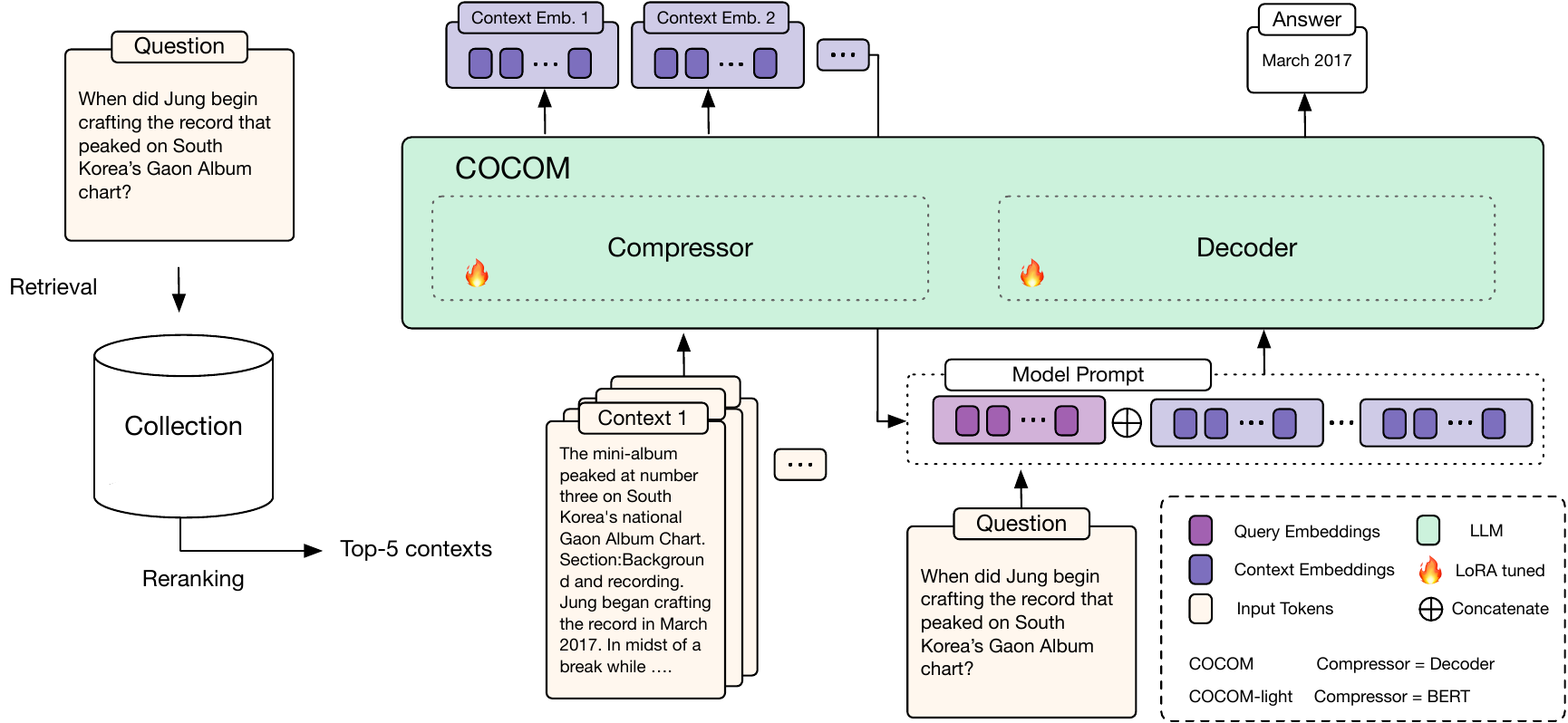}
    \caption{Overview of our \acrs(-light) model pipeline. }
    \label{fig:architecture}
\end{figure*}


The main idea of \acrs~is to enhance efficiency by compressing the context, which is typically given in surface form as input tokens into a smaller set of \textbf{context embeddings} which then serve as the input to the LLM. An overview of our entire pipeline is given in Figure \ref{fig:architecture}.
More formally, our approach can be described as follows:

Given a \textbf{context} $\mathcal{C}$ tokenized into a a sequence of  tokens $\{t_1, t_2,$ $\dots, t_n\}$, a compressor model $\phi_{comp}$, we compress $\mathcal{C}$ into
\textbf{context embeddings} $\mathcal{E}$, a smaller set of embeddings $\{e_1, e_2, \ldots, e_k\}$, where $k \ll n$. Each embedding $e_i \in \mathbb{R}^d$, with $d$ being the LLM's hidden dimension.
\begin{equation}
\label{eq:compression}
   \phi_{comp}: \{t_1, t_2, \ldots, t_n\} \rightarrow \{e_1, e_2, \ldots, e_k\} \in \mathbb{R}^d
\end{equation}

 Next, based on the compressed context embeddings  $\mathcal{E}$ and the user input $x$  the LLM $\phi_{LLM}$ generates a response $r$:
 \begin{equation}
      \theta_{LLM}: \{\mathcal{E},x\} \rightarrow r
 \end{equation}
The $\phi_{comp}$ model is trained to generate context embeddings that capture the content of the input tokens in a compressed form. As both models are trained jointly, $\theta_{LLM}$ learns to decode these context embeddings, extracting the relevant information required to answer user queries. 

\acrs~compresses the context-embeddings \emph{question independently}. This means not only do individual contexts have to be contextualized by an LLM only \emph{once}, but they can also be pre-computed offline and stored, drastically reducing computational costs of the LLM at inference time. Further, by only feeding a small number of context embeddings instead of the long context, the input size is reduced to a fraction leading to a massive speed-up for answer generation.


For \acrs, we utilize the same model for compression and answer generation $\phi_{comp} = \theta_{LLM}$. Therefore, we effectively train a single model on the two tasks. For the compression task, we prepend a special token \texttt{<AE>} to the input and depending on $\xi$ append a different number of context embedding tokens \texttt{<CTX>} at the end of the sequence. We directly use the representations of the last hidden layer as our context embeddings as input - to the same model - for the answer generation.  

As demonstrated later in the experiments, our method also allows us to potentially employ any embedding model as a compressor; including more lightweight encoder-only models such as BERT~\footnote{See Section~\ref{sec:cocom-light}}. 
\subsubsection{Adaptable Compression Rate}
The number of context embeddings $k = |\mathcal{E}|$ can be varied and allows to control the level of compression of the original context $\mathcal{C} = \{t_1, \dots, t_n\}$. We calculate the number of context embeddings $\xi$ per context $\mathcal{C}$ based on a compression rate $\xi$, and the length of the tokenized input $n = |\mathcal{C}|$.
\begin{equation}
    \xi =
    \left\lfloor \frac{n}{\xi} \right\rfloor \
\end{equation}
For instance, when compressing a context with length $n = 128$ with a compression rate $\xi = 64$ we obtain 2 context embeddings, reducing the input by 64 times.

\subsubsection{Multiple Contexts}
Knowledge-intensive tasks can benefit from providing the context of multiple retrieved passages \cite{izacard_atlas_2022, hsia2024ragged}, especially where reasoning over multiple contexts is necessary to solve the task \cite{fan-etal-2019-eli5, joshietal2017-triviaqa, yang-etal-2018-hotpotqa}. In classical RAG the contexts of multiple passages are concatenated and provided to the model. Similarly in \acrs~ we can provide \emph{context embeddings} of multiple passages to the LLM.  Contexts are compressed independently following Equation \ref{eq:compression}. We add [SEP] special tokens between the context embeddings before feeding them to the LLM to distinguish context stemming from different passages in the input.

\bigskip



\subsection{Pre-training Context Embeddings}
We propose two auto-regressive variations of the next-token prediction task to learn to compress context into context embeddings and to use these context embeddings as input to the LLM.

Following our earlier notation, the objective function for the standard next token prediction for input $\mathcal{X} = \{x_1, x_2, \dots, x_T\}$ can be written as:

\begin{equation}
  \mathcal{L}(\theta_{LLM}) = - \sum_{x_t \in \mathcal{X}} \log P_{\theta_{LLM}}(x_t \mid x_1, x_2, \ldots, x_{t-1})
\end{equation}

\subsubsection{Auto-encoding with Context Embeddings.} 
We modify the next token prediction task to recover the original input tokens from the compressed \emph{context embeddings} $\mathcal{E}$. This way we jointly train the compressor and LLM to decompress the original input which can be seen as a form of auto-encoding.  
\begin{equation}
\mathcal{E} = \phi_{comp}(x_1, x_2, \dots, x_T) 
\end{equation}

\begin{equation}
    \mathcal{L}(\theta_{LLM}, \phi_{comp}) = - \sum_{x_t \in \mathcal{X}} \log P_{\theta_{LLM}}(x_t \mid \mathcal{E}, x_1, \dots, x_{t-1})  
\end{equation}
This task serves as a preliminary step toward our final objective of answering questions from context embeddings. For this objective, we first aim to learn to compress and decompress the same input effectively. 
\subsubsection{Language Modeling from Context Embeddings.}
\label{sec:language_modeling}

Our final task is to answer questions based on the context embeddings. To this end, in our language modeling task, we train the model to continue a given input conditioned on context embeddings. 
This way the model learns not only to compress a given input but also to leverage the content of the context embeddings effectively. 

We split input $\mathcal{X} = \{ x_1, x_2, \dots, x_T\}$ into $\mathcal{X}_A = \{x_1, x_2, x_j\}$ and $\mathcal{X}_B = \{x_{j+1}, \dots, x_T\}$. After compressing the first part $\mathcal{X}_A$ into $\mathcal{E}_A$ we learn to generate the continuation - namely the second part $\mathcal{X}_B$ - based on the compressed representations $\mathcal{E}_A = \phi_{comp}(\mathcal{X}_A)$. This can be seen as a variation of the next token prediction task but conditioned on context embeddings.

\begin{equation}
    \mathcal{L}(\theta_{LLM}, \phi_{comp}) = - \sum_{x_t \in \mathcal{X}_B} \log P_{\theta_{LLM}}\big(x_t \mid \phi_{comp}(\mathcal{X}_A), x_1, \dots, x_{t-1}\big) 
\end{equation}
This language modeling task is complementary to the auto-encoding task. If we would only employ the auto-encoding from context embeddings task the LLM would be biased towards only recovering the original input, instead of leveraging the content of the context embeddings.
 
\subsection{Fine-tuning}
For the downstream RAG application, we fine-tune the model on a question $q$, relevant context(s) retrieved by a retrieval system and compressed into context embeddings  $\mathcal{E}$, which are combined into an instruction $I_{q, \mathcal{E}}$. We train the LLM to generate the target response $R = (r_1, r_2, \dots, t_T)$. We fine-tune our models on a combined set of publicly available QA datasets. 
We employ instruction fine-tuning only updating the models based on the target responses. 

\begin{equation}
    \mathcal{L}(\theta_{LLM}, \phi_{comp}) = - \sum_{r_t \in R} \log P_{\theta_{LLM}}(r_t \mid I_{ \mathcal{E}, q}, r_1, r_2, \dots, r_{t-1})
\end{equation}

\section{Experimental Setup}
\label{sec:exp}

In this section, we detail our experimental setup in terms of the RAG models and the five QA tasks.

\subsection{Implementation Details}
\label{sec:implementation_details}
We use \texttt{Mistral-7B-Instruct-v0.2}\footnote{\url{https://https://huggingface.co/mistralai/Mistral-7B-Instruct-v0.2}} as our backbone LLM for answer generation. For context compression in \acrs, we utilize the same model. For our more light-weight context compression, in \acrs-light, we employ
\texttt{bert-base-uncased}\footnote{\url{https://https://huggingface.co/google/bert-base-uncased}}. We apply three different compression rates: $\xi = {1, 16, 128}$.
We employ SPLADE-v3~\cite{lassance2024splade} with reranking top-50 using DeBERTa-v3~\cite{he2021debertav3} as our retrieval system. For all our experiments we use top-5 documents as context.
We release our strongest model checkpoints on Huggingface\footnote{\url{https://huggingface.co/collections/naver/cocom-6707e2b57e9dfd35279da238}.}.

\subsection{Training} 


For both pre-training and fine-tuning, we apply parameter-efficient LoRA tuning.
\subsubsection{Pre-training}
\label{sec:exp_pretraining}

For our pre-training, we employ the two earlier-mentioned pre-training autoencoding and language modeling tasks. Samples are drawn randomly with equal probability from both tasks. We tried different ratios but found this to perform best.
to ensure efficient batch processing, which requires that every sample in a batch contains a fixed-length tokenized input. To achieve this, we split the Wikipedia-KILT~\cite{petroni2020kilt} corpus \footnote{We publish this resource as a Huggingface dataset under \url{https://huggingface.co/datasets/dmrau/kilt-128}.} into chunks of 128 tokens using the Llama-2-7b tokenizer. 
We pre-train on in total 10m samples.
Training hyperparameters can be found in the Appendix in Table \ref{tab:pt_params}.

\subsubsection{Fine-tuning}
The BERGEN library~\cite{rau2024bergenbenchmarkinglibraryretrievalaugmented} is used to fine-tune the model. We fine-tune our models on various datasets concurrently. To construct our fine-tuning dataset~\footnote{We publish the dataset under \url{https://huggingface.co/datasets/dmrau/multi_qa}}, we combine training samples from Natural Questions~\cite{kwiatkowski2019natural},  MS MARCO~\footnote{We select only the first 100k queries.}~\cite{nguyen2016ms}, adversarial QA~\cite{bartolo2020beat},  HotpotQA~\cite{yang-etal-2018-hotpotqa},  \textit{WikiQA}~\cite{yang-etal-2015-wikiqa}, SCIQ~\cite{SciQ}, ASQA~\cite{stelmakh-etal-2022-asqa}, TriviaQA~\cite{joshietal2017-triviaqa}, Freebase QA~\cite{jiang-etal-2019-freebaseqa} and squad~\cite{rajpurkar-etal-2016-squad} - all of which are for question answering. Then we filter out queries with more than 128 tokens and labels of more than 64 tokens. For mode details we refer to Table \ref{tab:multi_dataset} in the Appendix.
Training hyperparameters can be found in the Appendix in Table \ref{tab:ft_params}.
\subsection{Evaluation}
We evaluate our model on several widely used QA datasets. Natural Questions \cite{kwiatkowski2019natural}, TriviaQA \cite{joshietal2017-triviaqa}, HotpotQA \cite{yang-etal-2018-hotpotqa}, ASQA \cite{stelmakh-etal-2022-asqa}, and PopQA \cite{mallen-etal-2023-trust}.

\subsubsection{Metrics}

As our main metric, following the standard protocol to evaluate fine-tuned models we use Exact Match (EM). To compare our results to previous works, which partially rely on untuned decoders and therefore produce verbose answers, we revert to the Match metric (M), which indicates whether the label is contained (as an exact match) in the generated answer. 



\subsection{Baselines without Context Compression}
We fine-tune the base model (Mistral-7B-Instruct-v0.2):

\begin{itemize}
    \item \textbf{RAG} - upper bound. The model receives the top-5 retrieved contexts, alongside the query and answers the question. This model serves as an upper bound in our experiment not applying context compression. 
    
    \item \textbf{Closed Book} - lower bound. (w/o RAG). The LLM generates an answer based on the query without any provided context. This serves as a lower-bound baseline.

\end{itemize}

\subsection{Baselines with Context Compression}

We compare our models to the context compression methods mentioned below. As mentioned earlier these models tune only parts of their model components on the downstream data but leave their decoder LLM untuned applying it zero-shot. We argue this to be a major limitation, as answering questions from context embeddings differs fundamentally from the standard language modeling hindering the model to effectively leverage the context embeddings.

To ensure comparability among approaches we use the same retrieval system as mentioned earlier in Section \ref{sec:implementation_details}.

\begin{itemize}
    \item \textbf{Autocompressor} ~\cite{chevalier2023adapting}: We use the \textit{princeton-nlp/AutoCompressor-Llama-2-7b-6k} checkpoint producing 50 summary vectors. As their model is limited to  compressing one single context, we just use the top retrieved document as context.

\item \textbf{ICAE}. ~\cite{ge2024incontext}: We use the Mistral-7B-Instruct-v0.2 LoRa-checkpoint\footnote{\url{https://huggingface.co/sggetao/icae}} which uses the same base LLM as ours and is therefore directly comparable. ICAE is fine-tuned to compress a single long context, however, in our work we use multiple contexts. To alleviate this we concatenate the top five retrieved contexts together as the context input for the model and truncate as the maximum length of 512 tokens. Note the model has a maximum output length of 128 compressed tokens, which approximately indicates a compression rate of 4 from its original concatenated context input.

\item \textbf{xRAG}. We utilize the xRAG-7b\footnote{\url{https://huggingface.co/Hannibal046/xrag-7b}}, and 8x7B mixture-of-experts model \footnote{https://huggingface.co/mistralai/Mixtral-8x7B-Instruct-v0.1}  alongside their strongest SFR compressor. The base model is again the same as ours, to ensure comparability. As their model is limited to compressing only a single context into a single compressed representation, we use the top retrieved context for the xRAG setting.\footnote{We also tested compressing five contexts together, which yielded lower effectiveness.} Again, their compressed representations stem from their dense retriever and are only adapted to the task through a simple linear projection which might limit the model its ability to compress contexts effectively.
We apply their predefined stopping criteria for answer generation, which aims at cutting the verbose nature of a untuned decoder LLM.
\end{itemize}

\begin{table*}[!ht]
    \caption{Results in Exact Match (EM) comparing \acrs(-light) to other context compression works. For Match metric (M) see Table \ref{tab:main_results_match} in Appendix. All methods use 5 context passages unless indicated otherwise. $^{\bigstar}$ Method limited to single context. $^{\bigtriangleup}$ upper baseline. $^{\bigtriangledown}$ lower baseline. $^*$ indicates statistical non-significance (p>0.05) with respect to COCOM $\xi$=4.}
    \label{tab:main_results_exact_match}
    \centering
    \setlength\tabcolsep{0pt} 
    \begin{tabular*}{\linewidth}{@{\extracolsep{\fill}} cp{0pt}llr*{5}{c}}
    \toprule
   \multicolumn{2}{c}{\bfseries Decoder} & \bfseries Method & \bf Compression rate ($\xi$) & \multicolumn{6}{c}{\bf Dataset} \\ 
    \cmidrule(lr){5-10}
    & & & & \bfseries NQ & \bfseries TriviaQA & \bfseries HotpotQA & \bfseries ASQA & \bfseries PopQA & \bfseries Average\\
    \midrule
   \multirow{5}{*}{\rotatebox{90}{Zero-shot}}& & AutoCompressor \cite{chevalier2023adapting}$^{\bigstar}$& $\times$ 4 & 0.000 &0.000 &0.000  &0.000 & 0.000 & 0.000 \\
   & & ICAE \cite{ge2024incontext}&   $\times$ 4 &0.210 & 0.592 & 0.184& 0.222 & 0.290 &0.300  \\
    & & xRAG \cite{cheng2024xrag}$^{\bigstar}$& \\
    & & \multicolumn{1}{r}{ \textit{Mistral-7B-v0.2}} &  $\times$ 128 &0.184 & 0.622 & 0.185 & 0.182 &  0.199 & 0.274 \\
   & &\multicolumn{1}{r}{\textit{Mixtral-8x7b}} &  $\times$ 128  & 0.265 & 0.744 & 0.239 & 0.292 & 0.318 & 0.372\\
    \midrule

    \multirow{9}{*}{\rotatebox{90}{Fine-tuned}}& \multirow{9}{*}{\rotatebox{90}{\textit{Mistral-7B-v0.2}}} & RAG$^{\bigtriangleup}$ (no compression) & -  & 0.597 & 0.883 & 0.500 & 0.622* & 0.514 & 0.623 \\
    &  & LLM$^{\bigtriangledown}$ (without context) & -  &0.359 & 0.708 & 0.264 & 0.546 & 0.199 & 0.416 \\ \cmidrule{3-10} 
   & &  \acrs-light \textit{(ours)}& $\times$ 4 & 0.539 & 0.849 & 0.409 & 0.601* & 0.458 & 0.531 \\
   & &&$\times$ 16 & 0.492 & 0.823 & 0.367 & 0.565 & 0.385 & 0.526 \\ 
   & &&$\times$ 128 & 0.444 & 0.794 & 0.321 & 0.550 & 0.314 & 0.485 \\ \cmidrule{3-10} 
   & & \acrs \textit{(ours)}  & $\times$ 4 & 0.554 & 0.859 & 0.430 & 0.609 & 0.474 & 0.585 \\
   & &&  $\times$ 16 & 0.539 & 0.852* & 0.426* & 0.602* & 0.465 & 0.577 \\
  & & &  $\times$ 128 & 0.511 & 0.835 & 0.378 & 0.585* & 0.391 & 0.540 \\
    \bottomrule
    \end{tabular*}
\end{table*}

\section{results}
\label{sec:res}

In this section, we present the main \acrs  and \acrs-light results in terms of effectiveness and efficiency.

\subsection{Main Results}
\label{sec:main_results}

The main results for \acrs are presented in Table~\ref{tab:main_results_exact_match}. We measure performance following the standard practice for fine-tuned models using the Exact Match (EM) metric.   
Compared to existing context compression methods\footnote{Side note: As previously mentioned earlier in Section~\ref{sec:related_work}, existing context compression methods do not tune the decoder LLM and therefore compare their performance and make effectiveness claims against zero-shot baselines. However, we argue that tuning compression models while freezing the decoder LLM could not be considered zero-shot, as it involves tuning some parts of the model on the task data. This setting is akin to soft-prompt tuning \cite{cuconasu2024power, li-liang-2021-prefix}, where the compressor model effectively parameterizes the soft prompt. Consequently, the performance of these methods should be regarded as intermediate between zero-shot and full decoder tuning and should be compared against similar tuning settings, such as soft prompt tuning.
}, our approach demonstrates a significantly (Tested with paired t-test (p<0.05)). higher effectiveness across different compression rates for all datasets tested. \acrs~even outperforms the much stronger xRAG Mixtral-8x7B model by a large margin having 8 times more parameters than \acrs. The highest performance is observed at a low compression rate ($\xi$=4). Increasing the compression rate results in a slight performance decline, which we will analyze further in Section \ref{sec:anlysis_context_compr}.

Compared to our upper bound baseline RAG without compression, we reduce the context by up to 128 times while still maintaining relatively high performance on average over datasets. 

Performance decreases on average 4 points for our strongest model (\acrs~$\xi$ = 4) and 10 points for the highest compression rate (\acrs~$\xi$ = 128). Compared to the lower bound baseline LLM without provided context we gain up to 17 points, adding only a small number of additional context embeddings to the input.  

Note, while EM is a standard metric for evaluating tuned models, it might underestimate zero-shot decoder methods that do not adapt the decoder to generate answers. To address this, we also provide results using the Match metric in the appendix in Table \ref{tab:main_results_match}. Although models that do not tune their decoder achieve relatively higher performance when measured in Match, our method’s effectiveness compared to other methods still remains consistently significantly higher.

 Overall, considering the effectiveness and the efficiency gains from context compression (discussed further in Section \ref{sec:efficiency}), \acrs~shows a very favorable trade-off.


\subsection{\acrs-light}
\label{sec:cocom-light}

Even though context compression has to be done only once offline, using a very large LLM can be costly, especially in resource-constraint settings. To this end, we propose \acrs-light, a computationally efficient context compression model based on BERT as a context compressor.

To alleviate the dimensional mismatch between the bert-based compressor and the - typically larger - LLM, we learn a linear projection layer $\bm{W}^{\xi b \times d}$, where $\xi$ is the compression rate, $b$ is the hidden dimension of BERT, and $d$ the hidden dimension of the LLM. To obtain a set of Context Embeddings we leverage the last hidden representation of each input token. We simply split the hidden representations into blocks of length $\xi$ and project each block into a single Context Embedding.  This way, we learn a block-wise aggregation of the input representations that depending on the input length, and the compression rate $\xi$ yields a different number of Context Embeddings per input. Note that a similar approach is applied in xRAG, where a projection layer is used on the embedding vector to resolve the dimensional mismatch. However, we argue that compressing using a single vector embedding could significantly restrict the compression quality, especially when using lightweight encoder models such as BERT. This restriction can result in much lower effectiveness compared to using a larger embedding model~\cite{cheng2024xrag}.

We present the results in Table \ref{tab:main_results_exact_match} measured in EM. Results for Match can be again found in the appendix in Table \ref{tab:main_results_match}. We find that while being highly effective for small compression rates to drop considerably for the highest compression rate of $\xi$=128. \acrs-light, compared to other methods poses an effective alternative to it's bigger counterpart \acrs, in resource-constrained settings.

\subsection{Computational Efficiency}
\label{sec:efficiency}

\begin{table}[!t]
    \centering
    
    \small
    \caption{Decoding efficiency in generation Time, GPU Memory, and number of operations (GFLOPs) for \acrs(-light) on dataset NQ.
    $\xi$ the compression rate. Efficiency speedup compared against RAG (no compression) is indicated in brackets.
    }
    \setlength\tabcolsep{0pt} 
    \begin{tabular*}{\linewidth}{@{\extracolsep{\fill}} lclll}
    \toprule
     \bf Model & \bf $\xi$ & \bf Decoding Time &  \bf GPU Mem.  & \bf GFLOPs  \\
     \textit{Mistral-7b-v0.2}& & (ms) & (GB) & \\
     \midrule
     RAG (no compr.)  & - & 1064 &  18.1 & 25031  \\
     LLM (no context.)  & - &159  &  14.1 & 607  \\ \midrule
     \acrs(-light)  & 4 &  371 ($\times$ 2.87 )&  15.1 ($\times$  1.20) & 7016 \phantom{d}($\times$  \phantom{2}3.57) \\
      & 16 & 213 ($\times$ 5.00 ) & 14.4 ($\times$  1.29) & 2465 \phantom{d}($\times$  10.16)  \\
      & 128 &  187 ($\times$ 5.69 ) &  14.2 ($\times$  1.27)& 1138 \phantom{d}($\times$  22.00) \\

     \bottomrule
    \end{tabular*}

    \label{tab:efficiency}
\end{table}

\begin{table}[!t]
    \centering
        \caption{Compression efficiency and storage requirements. Compressing ~24m contexts using  on a single A100 80GB GPU.}
    \setlength\tabcolsep{0pt} 
    \begin{tabular*}{\linewidth}{@{\extracolsep{\fill}} lccc}
    \toprule
     \bf Compressor & \bf $\xi$ & \bf Time (h)& \bf Index size (TB)\\
     \midrule 
     \textit{\acrs} & 4 & 89 & 6.06\\ 
     & 16 & 77 & 1.51\\ 
     & 128 & 73 & 0.19\\ 
     \textit{\acrs-light} & 4 &1 & 6.06\\ 
     & 16 &1& 1.51\\ 
     & 128 &1& 0.19 \\ 
          \bottomrule
    \end{tabular*}

    \label{tab:compression}
\end{table}

We measure efficiency in answer generation time (ms), maximum GPU memory (GB), and number of operations per generated token (Giga FLOPs) using the torch profiler. We run the experiments on a single A100 40GB with a fixed batch size of 16\footnote{Maximum batch size that fits on GPU across models.}.  We load the model in half-precision and use the PyTorch inference mode. We discard the first warm-up batch from
the measurement and measure the bare forward pass of the model. Note decoding results are independent of the compressor, therefore \acrs~ and \acrs-light share efficiency results.

We show our efficiency results for answer generation in Table \ref{tab:efficiency} for different compression rates $\xi$ and compare them to RAG without context compression. Context compression with \acrs~reduces answer generation time , GPU memory, and the number of operations drastically with up to 5.69 $\times$ less inference time cost, 1.27 $\times$ GPU memory, and 22 $\times$ GFLOPs compared to no compression.

In addition, Table~\ref{tab:compression} presents the compression costs for all documents in the kilt-100w (~24m contexts) collection using \acrs-light models at various compression rates. \acrs-light models demonstrate significantly faster compression speeds compared to the COCOM model by employing a much computationally lighter compressing module (up to 89 $\times$). Index size varies inversely with compression rate: higher compression rates result in smaller index storage requirements. However, this trade-off leads to lower quality in answer generation, as shown in Table~\ref{tab:main_results_exact_match}.

\subsection{Ablations}
In the following section, we run additional ablation experiments for \acrs~and \acrs-light. Most results can be found in Table \ref{tab:ablation}. We report performance in Exact Match on two datasets (NQ and ASQA).

\subsubsection{Handling multiple contexts.}
In table \ref{tab:topk}, we compare the performance of \acrs~with 1 retrieved context ($k=1$) versus our default setup $k=5$. On both datasets and for all compression rates, we observe a substantial gain when using more contexts. Moreover, \acrs~with 1 retrieved context is still significantly better compared to  other baselines relying on single retrieved document (ICAE,  xRAG) in table \ref{tab:main_results_exact_match}.
 As a decoder model by design should be able to handle multiple context representations, we argue that fine-tuning the decoder is a simple yet necessary solution compared to existing works.

\begin{table}
\caption{ Impact of the number of provided contexts (k) on \acrs~ measured in EM on datasets NQ and ASQA.}
\centering
\setlength\tabcolsep{0pt} 
\begin{tabular*}{\linewidth}{@{\extracolsep{\fill}} l*{5}{c}} 
\toprule
\bf   Model& \bf  $\xi$& \multicolumn{2}{c}{\bfseries NQ}  &  \multicolumn{2}{c}{ \bfseries ASQA} \\ 
&   &\bf  k=1 & \bf k=5 & \bf k=1 & \bf k=5   \\ 

\midrule
\acrs & 4  &  0.499 & 0.554  & 0.558 &  0.609   \\ 
& 16   &  0.491& 0.539     &  0.541 &  0.602  \\
& 128 &   0.482  & 0.511     & 0.544 &   0.585   \\
\bottomrule
\end{tabular*}
\label{tab:topk}
\end{table}

\subsubsection{Pre-training Context Compression}
Central to our approach is the compression of context into a small number of Context Embeddings. We argue that context compression fundamentally differs from the language modeling objective on which the model was originally trained. Consequently, we have employed auto-encoding and language-modeling-from-context-embedding tasks to learn how to effectively compress the context and utilize these compressed representations during decoding. We show the results of the impact of the pre-training tasks on the downstream performance after fine-tuning. Our results suggest that the dedicated pre-training tasks for context compression can improve performance for downstream QA performance, suggesting two possible explanations.  Either context compression is too complex to be learned concurrently with the downstream task, or larger fine-tuning datasets are necessary to effectively learn how to compress contexts.

\subsubsection{Pre-training Corpus}
Our method employs an initial pre-training step aimed at initializing context compression.  We train auto-regressively on the same target corpus, which is later used to retrieve relevant contexts.  In this experiment, our objective is to assess how variations in the pre-training corpus impact downstream QA performance, thereby testing the robustness of our approach. To explore this, we additionally pre-train the model on the "sample-10BT" subset of Fine-Web \cite{penedo2024finewebdatasetsdecantingweb}. We employ the same training methodology described in Section \ref{sec:exp_pretraining}, where we segment the collection into non-overlapping passages of 128 tokens using the Llama-2-7b tokenizer and train on a subset of 10 million tokens, similar to the target corpus.
The results presented in Table~\ref{tab:ablation} indicate a slight decrease in performance when using a different target corpus for pre-training. Nonetheless, our approach demonstrates robustness in handling variations in the pre-training corpus, highlighting its adaptability and effectiveness in context compression.

\subsubsection{ Decoder LLM Tuning}
Existing context compression methods tune only the compression module while keeping the decoder, responsible for generating the answer, frozen. A core distinction from these methods is that we tune all components including the decoder, in \acrs. We hypothesize that\textit{ context embeddings} differ significantly from the input token embeddings the model was trained on, thereby hindering effective utilization without dedicated tuning.
We investigate the consequences of freezing the decoder and solely tuning the compressor, akin to existing methods. Our findings show the criticality of tuning the decoder to achieve high effectiveness. This reinforces our hypothesis that specific tuning of \textit{context embeddings} seem essential for better performances.

\begin{table}[!t]
\caption{Impact of pre-training corpus, pre-training, and decoder tuning on downstream performance (EM). Compression rate $\xi$ = 128}
\centering
\setlength\tabcolsep{0pt} 
\begin{tabular*}{\linewidth}{@{\extracolsep{\fill}} lcc} 
\toprule
\bf Ablation & \multicolumn{2}{c}{\bfseries Datasets}\\ \cmidrule(lr){2-3}
&   \bf NQ &  \bf ASQA \\ 
\midrule
\acrs-light (baseline) &  0.444 & 0.550  \\
\multicolumn{1}{r}{w/o pre-training} & 0.423 & 0.524 \\
\multicolumn{1}{r}{pre-training on FineWeb} & 0.427 &0.545  \\
\multicolumn{1}{r}{w/o tuning decoder} & 0.353 & 0.438 \\
\midrule
\acrs (baseline) & 0.519 & 0.585\\
\multicolumn{1}{r}{w/o pre-training} & 0.490& 0.565 \\
\multicolumn{1}{r}{pre-training on FineWeb} & 0.503&0.581 \\
\multicolumn{1}{r}{w/o tuning decoder} & 0.421 & 0.521 \\

\bottomrule
\end{tabular*}
\label{tab:ablation}
\end{table}




\subsubsection{Fine-tuning Data}
\begin{figure}[!t]
    \centering
    \includegraphics[width=\linewidth]{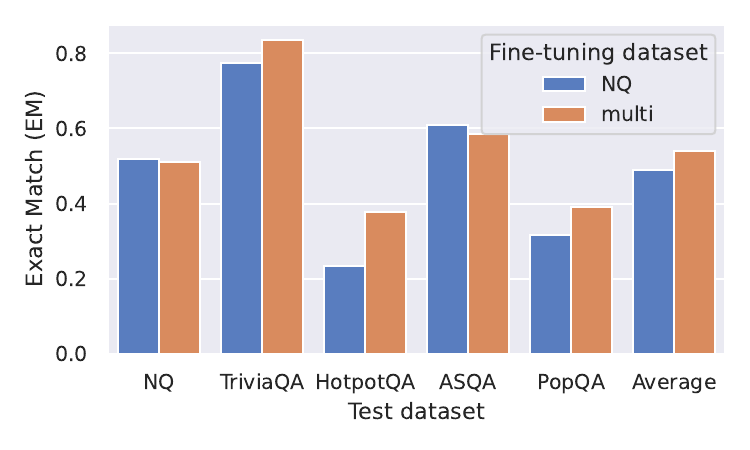}
    \caption{Impact on zero-shot transferability of fine-tuning on multiple datasets (multi) concurrently vs. on a single dataset for \acrs. Compression rate $\xi$ = 128}
    \label{fig:ft_dataset}
\end{figure}

In our experiments, we fine-tune our models simultaneously on multiple QA datasets before evaluating them on individual datasets. We explore the impact of this multi-dataset fine-tuning compared to training on a single dataset. Specifically, we fine-tune and evaluate our models on NQ (Natural Questions). For assessing transferability, we also conduct zero-shot evaluations on other datasets. The results are presented in Figure~\ref{fig:ft_dataset}. We find that fine-tuning solely on a single dataset, such as NQ, leads to slightly higher performance on that specific dataset. However, training on multiple datasets demonstrates superior transferability across all datasets, resulting in better average performance overall.

\section{Analysis}
\label{sec:ana}

In this section, we conduct further analysis on how compression affects the model.

\subsection{Context compression}
\label{sec:anlysis_context_compr}
In our earlier results in Section \ref{sec:main_results}, we observe a decline in performance with higher compression rates, particularly for the lightweight compressor in \acrs-light. To investigate potential reasons for this drop, we assess the model's ability to perform the two pre-training tasks: (i) compressing and decompressing input (auto-encoding) and (ii) language modeling from compressed representations after pre-training.

Table~\ref{tab:pretraining} showcases the results of these evaluations. Both the full and lightweight models effectively master the auto-encoding task at lower compression factors ($\xi = 4, 16$). However, they exhibit significant difficulties in reconstructing the input when the compression ratio increases ($\xi = 128$). This problem is notably more pronounced in our decoder-based compression model (COCOM).

We identify two possible explanations: First, compressing longer contexts into fewer embeddings inherently presents a challenge due to the inevitable information loss at higher compression rates. Second, the dimension of linear projection layers in the COCOM-light model is dependent on the compression rate; thus, a higher compression rate results in an increased parameter count within its linear layer to manage context compression. In contrast, the COCOM model employs \textit{lora} tuning, where the size of the components is not dependent on the compression rate. This fundamental difference in handling compression may explain why the COCOM-light model could potentially achieve higher effectiveness under conditions of high compression, due to its higher parameter count.
In terms of the second pre-training task, our results indicate that COCOM consistently outperforms COCOM-light, this finding also correlates to the final effectiveness of question-answering tasks, as indicated in table~\ref{tab:main_results_exact_match}.



\begin{table}[!t]
\caption{Pre-training evaluation on the tasks Auto Encoding (AE) and Language Modeling from Context Embeddings (LMCE) measured in Rouge-L.}
\centering
\setlength\tabcolsep{0pt} 
\begin{tabular*}{\linewidth}{@{\extracolsep{\fill}} l*{3}{c}} 
\toprule
\bf   Model& \bf  $\xi$& \multicolumn{2}{c}{\bfseries Rouge-L} \\ \cmidrule(lr){3-4}
&  & \bf AE &\bf  LMCE  \\  

 \midrule
\acrs-light & 4  & 0.9734 &0.1882 \\ 
& 16   & 0.9643 &0.1800 \\
& 128 & 0.7938 &0.1618 \\
\midrule
\acrs & 4   & 0.9979 &0.2045 \\
& 16   &0.9912 &0.1991 \\ 
 & 128   &0.5545 &0.1771 \\
\bottomrule
\end{tabular*}
\label{tab:pretraining}
\end{table}

\subsection{Case Study Answer Quality}

We investigate the answers generated with different models. For this, we randomly select a query from the NQ dataset and compare the responses generated by each method. Table~\ref{tab:case-study} presents the responses to the selected question.

From the responses, we observe that without RAG, the LLM tends to hallucinate and provide an irrelevant name as an answer. On the other hand, XRAG understands the question but returns an incorrect named entity, likely due to limitations in reading compressed embeddings accurately. ICAE struggles to comprehend the question, resulting in an unreasonable answer. Both COCOM and COCOM-light successfully answer the question correctly at a compression rate of 4. However, they encounter difficulties when the compression rate is increased to 128.

It is also worth noting that the XRAG response was intentionally truncated to a maximum of 30 tokens in its original publication, with the stopping criteria involving halting at punctuation mark such as periods, commas, and colons.

\begin{table}[h!]
\centering
\caption{Case Study: Generated responses using different methods. Dataset: NQ}
\scriptsize
\setlength\tabcolsep{0pt} 

\begin{tabularx}{\linewidth}{X}

\toprule
\multicolumn{1}{c}{\bf Model Input} \\
\midrule
\textbf{Question}: who played sarah hedley in when the boat comes in? \\[2pt]

\textbf{Context 1}: Rosalind Bailey. Rosalind Bailey Rosalind Bailey (born 1946) is a British actress, known for her portrayal of Sarah Headley ("née" Lytton) in the 1970s and 1980s BBC television drama "When the Boat Comes In". Bailey has appeared in numerous British television drama series, including "Byker Grove", "Distant Shores" and "Burn Up". Her stage work includes playing Miss Mary Shepherd in Alan Bennett’s play "The Lady in the Van". \\[2pt]
 
\textbf{Context 2}: Malcolm Terris. Malcolm Terris Malcolm Terris (born 11 January 1941 in Sunderland, County Durham) is a British actor. He had a lengthy career in a large number of television programmes. Possibly his best-known role was in "When the Boat Comes In", a popular 1970s series, where he played the part of Matt Headley. His film career includes appearances in "The First Great Train Robbery" (1978), "McVicar" (1980), "The Plague Dogs" (1982, voice only), "Slayground" (1983), "The Bounty" (1984) as Thomas Huggan, ship's surgeon, "Mata Hari" (1985), "Revolution" (1985), "Scandal" (1989), and "Chaplin" (1992). His TV appearances include: One episode of \\[2pt]

\textbf{Context 3}: When the Boat Comes In. When the Boat Comes In When the Boat Comes In is a British television period drama produced by the BBC between 1976 and 1981. The series stars James Bolam as Jack Ford, a First World War veteran who returns to his poverty-stricken (fictional) town of Gallowshield in the North East of England. The series dramatises the political struggles of the 1920s and 1930s and explores the impact of national and international politics upon Ford and the people around him. Section:Production. The majority of episodes were written by creator James Mitchell, but in Series 1 north-eastern \\[2pt]

\textbf{Context 4}: Susie Youssef. Youssef began her comedy career as a writer for "The Ronnie Johns Half Hour" in 2006, and made her acting debut in the short film "Clicked" in the role of Lina in 2011. In 2014, she played Jane in the short film "Kevin Needs to Make New Friends: Because Everyone Hates Him for Some Reason" and then turned to television where she appeared in "The Chaser's Media Circus". In 2014, Youssef played the lead role of Sarah in the Hayloft Project’s stage play "The Boat People" which won the Best On Stage award at the FBi SMAC Awards\\[2pt]

\textbf{Context 5}: Madelaine Newton. Madelaine Newton Madelaine Newton is a British actress best known for her portrayal of Dolly in 1970s BBC television drama "When the Boat Comes In". She is married to actor Kevin Whately, known for his role as Robert "Robbie" Lewis in both "Inspector Morse" and its spin-off "Lewis". They have two children. She starred alongside her husband in the "Inspector Morse" episode "Masonic Mysteries" as Beryl Newsome - the love-interest of Morse - whom Morse was wrongly suspected of murdering. She played Whately's on-screen wife in the 1988 Look and Read children's serial, Geordie Racer. She also made \\
\midrule
\multicolumn{1}{c}{\bf Generated Responses} \\
\midrule
\textbf{Label}:  
Rosalind Bailey \\[5pt] 
\textbf{LLM}: Anna Cropper \\[2pt]
\textbf{RAG}: Rosalind Bailey \\[2pt]
\textbf{xRAG}: 1976 \textcolor{gray}{: The role of Sarah Hedley in When the Boat Comes In was played by Rosalie Crutchley.}\\[2pt]
\textbf{ICAE Response}: Sarah Hadland \\[2pt]
\textbf{\acrs-4}: Rosalind Bailey \\[2pt]
\textbf{\acrs-light-4}: Rosalind Bailey \\[2pt]
\textbf{\acrs-128}: Alison Steadman \\[2pt]
\textbf{\acrs-light-128}: Rosalind Elliott \\
\bottomrule
\end{tabularx}
\label{tab:case-study}
\end{table}

\section{Conclusion}
\label{sec:con}

In this paper, we presented our novel approach \acrs approach for context compression.
Our main finding is that 
\acrs accelerates answer generation, by reducing the model's input, by compressing multiple contexts into context embeddings that, once pre-computed serve to augment the answer generation.

Our approach maximizes the potential of the LLM by tuning all components outperforming existing methods for context compression in RAG. By offering a trade-off between efficiency and effectiveness, our method allows for the selection of varying numbers of context compression tokens. This flexibility enables us to balance higher answer quality against faster generation times as needed.
Unlike previous methods, our approach allows for the input of multiple contexts, which enhances generation quality and optimally makes use of the reduced decoding time. This is because only for very long inputs, the distinction between the context in token form and a reduced set of embeddings becomes most apparent. 

We hope that our work will inspire further research in context compression and pave the way for efficient and effective deployment of Retrieval-Augmented Generation (RAG) models in real-world applications.

\section{Limitations}
\label{sec:lim}
We end this paper by discussing the remaining  limitations of our model and of our experiments.

Our approach offers great potential to reduce the computational footprint of RAG.  
However, in our experiments we were constrained by computational resources, which limits us to utilizing a relatively small model of 7 billion parameters. This constraint prevents us from exploring the capabilities of larger models such as LLaMA70B or Mixtral7x8B, which might offer enhanced performance but require significant computational power for training and inference.

Our approach demonstrates the potential to leverage a much larger set of documents compared to non-compressed models, leading to notable efficiency gains. These gains are particularly evident when dealing with a substantial volume of documents. However, due to resource limitations, our experiments have been restricted to only 5 documents. This limited scope may not fully reflect the method’s effectiveness when scaled to larger document collections, where the benefits could be more pronounced.

Additionally, the evaluation of our method has been conducted exclusively on Question Answering (QA) tasks and using English corpora. A more comprehensive assessment, encompassing diverse tasks and multilingual datasets, would be necessary to thoroughly understand the model’s capabilities and limitations in different scenarios.






\bibliographystyle{ACM-Reference-Format}
\bibliography{bib}


\begin{thebibliography}{36}


\ifx \showCODEN    \undefined \def \showCODEN     #1{\unskip}     \fi
\ifx \showDOI      \undefined \def \showDOI       #1{#1}\fi
\ifx \showISBNx    \undefined \def \showISBNx     #1{\unskip}     \fi
\ifx \showISBNxiii \undefined \def \showISBNxiii  #1{\unskip}     \fi
\ifx \showISSN     \undefined \def \showISSN      #1{\unskip}     \fi
\ifx \showLCCN     \undefined \def \showLCCN      #1{\unskip}     \fi
\ifx \shownote     \undefined \def \shownote      #1{#1}          \fi
\ifx \showarticletitle \undefined \def \showarticletitle #1{#1}   \fi
\ifx \showURL      \undefined \def \showURL       {\relax}        \fi
\providecommand\bibfield[2]{#2}
\providecommand\bibinfo[2]{#2}
\providecommand\natexlab[1]{#1}
\providecommand\showeprint[2][]{arXiv:#2}

\bibitem[Asai et~al\mbox{.}(2024)]%
        {asai2024reliable}
\bibfield{author}{\bibinfo{person}{Akari Asai}, \bibinfo{person}{Zexuan Zhong}, \bibinfo{person}{Danqi Chen}, \bibinfo{person}{Pang~Wei Koh}, \bibinfo{person}{Luke Zettlemoyer}, \bibinfo{person}{Hannaneh Hajishirzi}, {and} \bibinfo{person}{Wen-tau Yih}.} \bibinfo{year}{2024}\natexlab{}.
\newblock \showarticletitle{Reliable, Adaptable, and Attributable Language Models with Retrieval}.
\newblock \bibinfo{journal}{\emph{arXiv preprint arXiv:2403.03187}} (\bibinfo{year}{2024}).
\newblock


\bibitem[Bartolo et~al\mbox{.}(2020)]%
        {bartolo2020beat}
\bibfield{author}{\bibinfo{person}{Max Bartolo}, \bibinfo{person}{Alastair Roberts}, \bibinfo{person}{Johannes Welbl}, \bibinfo{person}{Sebastian Riedel}, {and} \bibinfo{person}{Pontus Stenetorp}.} \bibinfo{year}{2020}\natexlab{}.
\newblock \showarticletitle{Beat the AI: Investigating Adversarial Human Annotation for Reading Comprehension}.
\newblock \bibinfo{journal}{\emph{Transactions of the Association for Computational Linguistics}}  \bibinfo{volume}{8} (\bibinfo{year}{2020}), \bibinfo{pages}{662--678}.
\newblock
\urldef\tempurl%
\url{https://doi.org/10.1162/tacl\_a\_00338}
\showDOI{\tempurl}
\showeprint{https://doi.org/10.1162/tacl\_a\_00338}


\bibitem[Cheng et~al\mbox{.}(2024)]%
        {cheng2024xrag}
\bibfield{author}{\bibinfo{person}{Xin Cheng}, \bibinfo{person}{Xun Wang}, \bibinfo{person}{Xingxing Zhang}, \bibinfo{person}{Tao Ge}, \bibinfo{person}{Si-Qing Chen}, \bibinfo{person}{Furu Wei}, \bibinfo{person}{Huishuai Zhang}, {and} \bibinfo{person}{Dongyan Zhao}.} \bibinfo{year}{2024}\natexlab{}.
\newblock \showarticletitle{xRAG: Extreme Context Compression for Retrieval-augmented Generation with One Token}.
\newblock \bibinfo{journal}{\emph{arXiv preprint arXiv:2405.13792}} (\bibinfo{year}{2024}).
\newblock


\bibitem[Chevalier et~al\mbox{.}(2023)]%
        {chevalier2023adapting}
\bibfield{author}{\bibinfo{person}{Alexis Chevalier}, \bibinfo{person}{Alexander Wettig}, \bibinfo{person}{Anirudh Ajith}, {and} \bibinfo{person}{Danqi Chen}.} \bibinfo{year}{2023}\natexlab{}.
\newblock \bibinfo{title}{Adapting Language Models to Compress Contexts}.
\newblock
\newblock
\showeprint[arxiv]{2305.14788}~[cs.CL]


\bibitem[Cuconasu et~al\mbox{.}(2024)]%
        {cuconasu2024power}
\bibfield{author}{\bibinfo{person}{Florin Cuconasu}, \bibinfo{person}{Giovanni Trappolini}, \bibinfo{person}{Federico Siciliano}, \bibinfo{person}{Simone Filice}, \bibinfo{person}{Cesare Campagnano}, \bibinfo{person}{Yoelle Maarek}, \bibinfo{person}{Nicola Tonellotto}, {and} \bibinfo{person}{Fabrizio Silvestri}.} \bibinfo{year}{2024}\natexlab{}.
\newblock \bibinfo{title}{The Power of Noise: Redefining Retrieval for RAG Systems}.
\newblock
\newblock
\showeprint[arxiv]{2401.14887}~[cs.IR]


\bibitem[Dehghani et~al\mbox{.}(2019)]%
        {DBLP:conf/wsdm/DehghaniAKR19}
\bibfield{author}{\bibinfo{person}{Mostafa Dehghani}, \bibinfo{person}{Hosein Azarbonyad}, \bibinfo{person}{Jaap Kamps}, {and} \bibinfo{person}{Maarten de Rijke}.} \bibinfo{year}{2019}\natexlab{}.
\newblock \showarticletitle{Learning to Transform, Combine, and Reason in Open-Domain Question Answering}. In \bibinfo{booktitle}{\emph{Proceedings of the Twelfth {ACM} International Conference on Web Search and Data Mining, {WSDM} 2019, Melbourne, VIC, Australia, February 11-15, 2019}}, \bibfield{editor}{\bibinfo{person}{J.~Shane Culpepper}, \bibinfo{person}{Alistair Moffat}, \bibinfo{person}{Paul~N. Bennett}, {and} \bibinfo{person}{Kristina Lerman}} (Eds.). \bibinfo{publisher}{{ACM}}, \bibinfo{pages}{681--689}.
\newblock
\urldef\tempurl%
\url{https://doi.org/10.1145/3289600.3291012}
\showDOI{\tempurl}


\bibitem[Fan et~al\mbox{.}(2019)]%
        {fan-etal-2019-eli5}
\bibfield{author}{\bibinfo{person}{Angela Fan}, \bibinfo{person}{Yacine Jernite}, \bibinfo{person}{Ethan Perez}, \bibinfo{person}{David Grangier}, \bibinfo{person}{Jason Weston}, {and} \bibinfo{person}{Michael Auli}.} \bibinfo{year}{2019}\natexlab{}.
\newblock \showarticletitle{{ELI}5: Long Form Question Answering}. In \bibinfo{booktitle}{\emph{Proceedings of the 57th Annual Meeting of the Association for Computational Linguistics}}, \bibfield{editor}{\bibinfo{person}{Anna Korhonen}, \bibinfo{person}{David Traum}, {and} \bibinfo{person}{Llu{\'\i}s M{\`a}rquez}} (Eds.). \bibinfo{publisher}{Association for Computational Linguistics}, \bibinfo{address}{Florence, Italy}, \bibinfo{pages}{3558--3567}.
\newblock
\urldef\tempurl%
\url{https://doi.org/10.18653/v1/P19-1346}
\showDOI{\tempurl}


\bibitem[Ge et~al\mbox{.}(2024)]%
        {ge2024incontext}
\bibfield{author}{\bibinfo{person}{Tao Ge}, \bibinfo{person}{Hu Jing}, \bibinfo{person}{Lei Wang}, \bibinfo{person}{Xun Wang}, \bibinfo{person}{Si-Qing Chen}, {and} \bibinfo{person}{Furu Wei}.} \bibinfo{year}{2024}\natexlab{}.
\newblock \showarticletitle{In-context Autoencoder for Context Compression in a Large Language Model}. In \bibinfo{booktitle}{\emph{The Twelfth International Conference on Learning Representations}}.
\newblock
\urldef\tempurl%
\url{https://openreview.net/forum?id=uREj4ZuGJE}
\showURL{%
\tempurl}


\bibitem[He et~al\mbox{.}(2021)]%
        {he2021debertav3}
\bibfield{author}{\bibinfo{person}{Pengcheng He}, \bibinfo{person}{Jianfeng Gao}, {and} \bibinfo{person}{Weizhu Chen}.} \bibinfo{year}{2021}\natexlab{}.
\newblock \showarticletitle{Debertav3: Improving deberta using electra-style pre-training with gradient-disentangled embedding sharing}.
\newblock \bibinfo{journal}{\emph{arXiv preprint arXiv:2111.09543}} (\bibinfo{year}{2021}).
\newblock


\bibitem[Hsia et~al\mbox{.}(2024)]%
        {hsia2024ragged}
\bibfield{author}{\bibinfo{person}{Jennifer Hsia}, \bibinfo{person}{Afreen Shaikh}, \bibinfo{person}{Zhiruo Wang}, {and} \bibinfo{person}{Graham Neubig}.} \bibinfo{year}{2024}\natexlab{}.
\newblock \bibinfo{title}{RAGGED: Towards Informed Design of Retrieval Augmented Generation Systems}.
\newblock
\newblock
\showeprint[arxiv]{2403.09040}~[cs.CL]


\bibitem[Izacard and Grave(2021)]%
        {izacard_leveraging_2021}
\bibfield{author}{\bibinfo{person}{Gautier Izacard} {and} \bibinfo{person}{Edouard Grave}.} \bibinfo{year}{2021}\natexlab{}.
\newblock \bibinfo{title}{Leveraging {Passage} {Retrieval} with {Generative} {Models} for {Open} {Domain} {Question} {Answering}}.
\newblock
\newblock
\urldef\tempurl%
\url{http://arxiv.org/abs/2007.01282}
\showURL{%
\tempurl}
\newblock
\shownote{arXiv:2007.01282 [cs]}.


\bibitem[Izacard et~al\mbox{.}(2022)]%
        {izacard_atlas_2022}
\bibfield{author}{\bibinfo{person}{Gautier Izacard}, \bibinfo{person}{Patrick Lewis}, \bibinfo{person}{Maria Lomeli}, \bibinfo{person}{Lucas Hosseini}, \bibinfo{person}{Fabio Petroni}, \bibinfo{person}{Timo Schick}, \bibinfo{person}{Jane Dwivedi-Yu}, \bibinfo{person}{Armand Joulin}, \bibinfo{person}{Sebastian Riedel}, {and} \bibinfo{person}{Edouard Grave}.} \bibinfo{year}{2022}\natexlab{}.
\newblock \bibinfo{title}{Atlas: {Few}-shot {Learning} with {Retrieval} {Augmented} {Language} {Models}}.
\newblock
\newblock
\urldef\tempurl%
\url{http://arxiv.org/abs/2208.03299}
\showURL{%
\tempurl}
\newblock
\shownote{arXiv:2208.03299 [cs]}.


\bibitem[Jiang et~al\mbox{.}(2023)]%
        {jiang-etal-2023-llmlingua}
\bibfield{author}{\bibinfo{person}{Huiqiang Jiang}, \bibinfo{person}{Qianhui Wu}, \bibinfo{person}{Chin-Yew Lin}, \bibinfo{person}{Yuqing Yang}, {and} \bibinfo{person}{Lili Qiu}.} \bibinfo{year}{2023}\natexlab{}.
\newblock \showarticletitle{{LLML}ingua: Compressing Prompts for Accelerated Inference of Large Language Models}. In \bibinfo{booktitle}{\emph{Proceedings of the 2023 Conference on Empirical Methods in Natural Language Processing}}, \bibfield{editor}{\bibinfo{person}{Houda Bouamor}, \bibinfo{person}{Juan Pino}, {and} \bibinfo{person}{Kalika Bali}} (Eds.). \bibinfo{publisher}{Association for Computational Linguistics}, \bibinfo{address}{Singapore}, \bibinfo{pages}{13358--13376}.
\newblock
\urldef\tempurl%
\url{https://doi.org/10.18653/v1/2023.emnlp-main.825}
\showDOI{\tempurl}


\bibitem[Jiang et~al\mbox{.}(2019)]%
        {jiang-etal-2019-freebaseqa}
\bibfield{author}{\bibinfo{person}{Kelvin Jiang}, \bibinfo{person}{Dekun Wu}, {and} \bibinfo{person}{Hui Jiang}.} \bibinfo{year}{2019}\natexlab{}.
\newblock \showarticletitle{{F}reebase{QA}: A New Factoid {QA} Data Set Matching Trivia-Style Question-Answer Pairs with {F}reebase}. In \bibinfo{booktitle}{\emph{Proceedings of the 2019 Conference of the North {A}merican Chapter of the Association for Computational Linguistics: Human Language Technologies, Volume 1 (Long and Short Papers)}}. \bibinfo{publisher}{Association for Computational Linguistics}, \bibinfo{address}{Minneapolis, Minnesota}, \bibinfo{pages}{318--323}.
\newblock
\urldef\tempurl%
\url{https://doi.org/10.18653/v1/N19-1028}
\showDOI{\tempurl}


\bibitem[Johannes~Welbl(2017)]%
        {SciQ}
\bibfield{author}{\bibinfo{person}{Matt~Gardner Johannes~Welbl, Nelson F.~Liu}.} \bibinfo{year}{2017}\natexlab{}.
\newblock \showarticletitle{Crowdsourcing Multiple Choice Science Questions}.
\newblock \bibinfo{journal}{\emph{arXiv:1707.06209v1}}.
\newblock


\bibitem[Joshi et~al\mbox{.}(2017)]%
        {joshietal2017-triviaqa}
\bibfield{author}{\bibinfo{person}{Mandar Joshi}, \bibinfo{person}{Eunsol Choi}, \bibinfo{person}{Daniel Weld}, {and} \bibinfo{person}{Luke Zettlemoyer}.} \bibinfo{year}{2017}\natexlab{}.
\newblock \showarticletitle{{T}rivia{QA}: A Large Scale Distantly Supervised Challenge Dataset for Reading Comprehension}. In \bibinfo{booktitle}{\emph{Proceedings of the 55th Annual Meeting of the Association for Computational Linguistics (Volume 1: Long Papers)}}, \bibfield{editor}{\bibinfo{person}{Regina Barzilay} {and} \bibinfo{person}{Min-Yen Kan}} (Eds.). \bibinfo{publisher}{Association for Computational Linguistics}, \bibinfo{address}{Vancouver, Canada}, \bibinfo{pages}{1601--1611}.
\newblock
\urldef\tempurl%
\url{https://doi.org/10.18653/v1/P17-1147}
\showDOI{\tempurl}


\bibitem[Kwiatkowski et~al\mbox{.}(2019)]%
        {kwiatkowski2019natural}
\bibfield{author}{\bibinfo{person}{Tom Kwiatkowski}, \bibinfo{person}{Jennimaria Palomaki}, \bibinfo{person}{Olivia Redfield}, \bibinfo{person}{Michael Collins}, \bibinfo{person}{Ankur Parikh}, \bibinfo{person}{Chris Alberti}, \bibinfo{person}{Danielle Epstein}, \bibinfo{person}{Illia Polosukhin}, \bibinfo{person}{Jacob Devlin}, \bibinfo{person}{Kenton Lee}, {et~al\mbox{.}}} \bibinfo{year}{2019}\natexlab{}.
\newblock \showarticletitle{Natural questions: a benchmark for question answering research}.
\newblock \bibinfo{journal}{\emph{Transactions of the Association for Computational Linguistics}}  \bibinfo{volume}{7} (\bibinfo{year}{2019}), \bibinfo{pages}{453--466}.
\newblock


\bibitem[Lassance et~al\mbox{.}(2024)]%
        {lassance2024splade}
\bibfield{author}{\bibinfo{person}{Carlos Lassance}, \bibinfo{person}{Herv{\'e} D{\'e}jean}, \bibinfo{person}{Thibault Formal}, {and} \bibinfo{person}{St{\'e}phane Clinchant}.} \bibinfo{year}{2024}\natexlab{}.
\newblock \showarticletitle{SPLADE-v3: New baselines for SPLADE}.
\newblock \bibinfo{journal}{\emph{arXiv preprint arXiv:2403.06789}} (\bibinfo{year}{2024}).
\newblock


\bibitem[Li and Liang(2021)]%
        {li-liang-2021-prefix}
\bibfield{author}{\bibinfo{person}{Xiang~Lisa Li} {and} \bibinfo{person}{Percy Liang}.} \bibinfo{year}{2021}\natexlab{}.
\newblock \showarticletitle{Prefix-Tuning: Optimizing Continuous Prompts for Generation}. In \bibinfo{booktitle}{\emph{Proceedings of the 59th Annual Meeting of the Association for Computational Linguistics and the 11th International Joint Conference on Natural Language Processing (Volume 1: Long Papers)}}, \bibfield{editor}{\bibinfo{person}{Chengqing Zong}, \bibinfo{person}{Fei Xia}, \bibinfo{person}{Wenjie Li}, {and} \bibinfo{person}{Roberto Navigli}} (Eds.). \bibinfo{publisher}{Association for Computational Linguistics}, \bibinfo{address}{Online}, \bibinfo{pages}{4582--4597}.
\newblock
\urldef\tempurl%
\url{https://doi.org/10.18653/v1/2021.acl-long.353}
\showDOI{\tempurl}


\bibitem[Li(2023)]%
        {li2023unlocking}
\bibfield{author}{\bibinfo{person}{Yucheng Li}.} \bibinfo{year}{2023}\natexlab{}.
\newblock \bibinfo{title}{Unlocking Context Constraints of LLMs: Enhancing Context Efficiency of LLMs with Self-Information-Based Content Filtering}.
\newblock
\newblock
\showeprint[arxiv]{2304.12102}~[cs.CL]


\bibitem[Liu et~al\mbox{.}(2023)]%
        {liu_lost_2023}
\bibfield{author}{\bibinfo{person}{Nelson~F. Liu}, \bibinfo{person}{Kevin Lin}, \bibinfo{person}{John Hewitt}, \bibinfo{person}{Ashwin Paranjape}, \bibinfo{person}{Michele Bevilacqua}, \bibinfo{person}{Fabio Petroni}, {and} \bibinfo{person}{Percy Liang}.} \bibinfo{year}{2023}\natexlab{}.
\newblock \bibinfo{title}{Lost in the {Middle}: {How} {Language} {Models} {Use} {Long} {Contexts}}.
\newblock
\newblock
\urldef\tempurl%
\url{https://doi.org/10.48550/arXiv.2307.03172}
\showDOI{\tempurl}
\newblock
\shownote{arXiv:2307.03172 [cs]}.


\bibitem[Mallen et~al\mbox{.}(2023)]%
        {mallen-etal-2023-trust}
\bibfield{author}{\bibinfo{person}{Alex Mallen}, \bibinfo{person}{Akari Asai}, \bibinfo{person}{Victor Zhong}, \bibinfo{person}{Rajarshi Das}, \bibinfo{person}{Daniel Khashabi}, {and} \bibinfo{person}{Hannaneh Hajishirzi}.} \bibinfo{year}{2023}\natexlab{}.
\newblock \showarticletitle{When Not to Trust Language Models: Investigating Effectiveness of Parametric and Non-Parametric Memories}. In \bibinfo{booktitle}{\emph{Proceedings of the 61st Annual Meeting of the Association for Computational Linguistics (Volume 1: Long Papers)}}, \bibfield{editor}{\bibinfo{person}{Anna Rogers}, \bibinfo{person}{Jordan Boyd-Graber}, {and} \bibinfo{person}{Naoaki Okazaki}} (Eds.). \bibinfo{publisher}{Association for Computational Linguistics}, \bibinfo{address}{Toronto, Canada}, \bibinfo{pages}{9802--9822}.
\newblock
\urldef\tempurl%
\url{https://doi.org/10.18653/v1/2023.acl-long.546}
\showDOI{\tempurl}


\bibitem[Morris et~al\mbox{.}(2023)]%
        {morris-etal-2023-text}
\bibfield{author}{\bibinfo{person}{John Morris}, \bibinfo{person}{Volodymyr Kuleshov}, \bibinfo{person}{Vitaly Shmatikov}, {and} \bibinfo{person}{Alexander Rush}.} \bibinfo{year}{2023}\natexlab{}.
\newblock \showarticletitle{Text Embeddings Reveal (Almost) As Much As Text}. In \bibinfo{booktitle}{\emph{Proceedings of the 2023 Conference on Empirical Methods in Natural Language Processing}}, \bibfield{editor}{\bibinfo{person}{Houda Bouamor}, \bibinfo{person}{Juan Pino}, {and} \bibinfo{person}{Kalika Bali}} (Eds.). \bibinfo{publisher}{Association for Computational Linguistics}, \bibinfo{address}{Singapore}, \bibinfo{pages}{12448--12460}.
\newblock
\urldef\tempurl%
\url{https://doi.org/10.18653/v1/2023.emnlp-main.765}
\showDOI{\tempurl}


\bibitem[Muennighoff et~al\mbox{.}(2024)]%
        {muennighoff2024generative}
\bibfield{author}{\bibinfo{person}{Niklas Muennighoff}, \bibinfo{person}{Hongjin Su}, \bibinfo{person}{Liang Wang}, \bibinfo{person}{Nan Yang}, \bibinfo{person}{Furu Wei}, \bibinfo{person}{Tao Yu}, \bibinfo{person}{Amanpreet Singh}, {and} \bibinfo{person}{Douwe Kiela}.} \bibinfo{year}{2024}\natexlab{}.
\newblock \bibinfo{title}{Generative Representational Instruction Tuning}.
\newblock
\newblock
\showeprint[arxiv]{2402.09906}~[cs.CL]


\bibitem[Nguyen et~al\mbox{.}(2016)]%
        {nguyen2016ms}
\bibfield{author}{\bibinfo{person}{Tri Nguyen}, \bibinfo{person}{Mir Rosenberg}, \bibinfo{person}{Xia Song}, \bibinfo{person}{Jianfeng Gao}, \bibinfo{person}{Saurabh Tiwary}, \bibinfo{person}{Rangan Majumder}, {and} \bibinfo{person}{Li Deng}.} \bibinfo{year}{2016}\natexlab{}.
\newblock \showarticletitle{Ms marco: A human-generated machine reading comprehension dataset}.
\newblock  (\bibinfo{year}{2016}).
\newblock


\bibitem[Penedo et~al\mbox{.}(2024)]%
        {penedo2024finewebdatasetsdecantingweb}
\bibfield{author}{\bibinfo{person}{Guilherme Penedo}, \bibinfo{person}{Hynek Kydlíček}, \bibinfo{person}{Loubna~Ben allal}, \bibinfo{person}{Anton Lozhkov}, \bibinfo{person}{Margaret Mitchell}, \bibinfo{person}{Colin Raffel}, \bibinfo{person}{Leandro~Von Werra}, {and} \bibinfo{person}{Thomas Wolf}.} \bibinfo{year}{2024}\natexlab{}.
\newblock \bibinfo{title}{The FineWeb Datasets: Decanting the Web for the Finest Text Data at Scale}.
\newblock
\newblock
\showeprint[arxiv]{2406.17557}~[cs.CL]
\urldef\tempurl%
\url{https://arxiv.org/abs/2406.17557}
\showURL{%
\tempurl}


\bibitem[Petroni et~al\mbox{.}(2020)]%
        {petroni2020kilt}
\bibfield{author}{\bibinfo{person}{Fabio Petroni}, \bibinfo{person}{Aleksandra Piktus}, \bibinfo{person}{Angela Fan}, \bibinfo{person}{Patrick Lewis}, \bibinfo{person}{Majid Yazdani}, \bibinfo{person}{Nicola De~Cao}, \bibinfo{person}{James Thorne}, \bibinfo{person}{Yacine Jernite}, \bibinfo{person}{Vladimir Karpukhin}, \bibinfo{person}{Jean Maillard}, {et~al\mbox{.}}} \bibinfo{year}{2020}\natexlab{}.
\newblock \showarticletitle{KILT: a benchmark for knowledge intensive language tasks}.
\newblock \bibinfo{journal}{\emph{arXiv preprint arXiv:2009.02252}} (\bibinfo{year}{2020}).
\newblock


\bibitem[Rajpurkar et~al\mbox{.}(2016)]%
        {rajpurkar-etal-2016-squad}
\bibfield{author}{\bibinfo{person}{Pranav Rajpurkar}, \bibinfo{person}{Jian Zhang}, \bibinfo{person}{Konstantin Lopyrev}, {and} \bibinfo{person}{Percy Liang}.} \bibinfo{year}{2016}\natexlab{}.
\newblock \showarticletitle{{SQ}u{AD}: 100,000+ Questions for Machine Comprehension of Text}. In \bibinfo{booktitle}{\emph{Proceedings of the 2016 Conference on Empirical Methods in Natural Language Processing}}, \bibfield{editor}{\bibinfo{person}{Jian Su}, \bibinfo{person}{Kevin Duh}, {and} \bibinfo{person}{Xavier Carreras}} (Eds.). \bibinfo{publisher}{Association for Computational Linguistics}, \bibinfo{address}{Austin, Texas}, \bibinfo{pages}{2383--2392}.
\newblock
\urldef\tempurl%
\url{https://doi.org/10.18653/v1/D16-1264}
\showDOI{\tempurl}
\showeprint[arxiv]{1606.05250}~[cs.CL]


\bibitem[Rau et~al\mbox{.}(2024)]%
        {rau2024bergenbenchmarkinglibraryretrievalaugmented}
\bibfield{author}{\bibinfo{person}{David Rau}, \bibinfo{person}{Hervé Déjean}, \bibinfo{person}{Nadezhda Chirkova}, \bibinfo{person}{Thibault Formal}, \bibinfo{person}{Shuai Wang}, \bibinfo{person}{Vassilina Nikoulina}, {and} \bibinfo{person}{Stéphane Clinchant}.} \bibinfo{year}{2024}\natexlab{}.
\newblock \bibinfo{title}{BERGEN: A Benchmarking Library for Retrieval-Augmented Generation}.
\newblock
\newblock
\showeprint[arxiv]{2407.01102}~[cs.CL]
\urldef\tempurl%
\url{https://arxiv.org/abs/2407.01102}
\showURL{%
\tempurl}


\bibitem[Stelmakh et~al\mbox{.}(2022)]%
        {stelmakh-etal-2022-asqa}
\bibfield{author}{\bibinfo{person}{Ivan Stelmakh}, \bibinfo{person}{Yi Luan}, \bibinfo{person}{Bhuwan Dhingra}, {and} \bibinfo{person}{Ming-Wei Chang}.} \bibinfo{year}{2022}\natexlab{}.
\newblock \showarticletitle{{ASQA}: Factoid Questions Meet Long-Form Answers}. In \bibinfo{booktitle}{\emph{Proceedings of the 2022 Conference on Empirical Methods in Natural Language Processing}}, \bibfield{editor}{\bibinfo{person}{Yoav Goldberg}, \bibinfo{person}{Zornitsa Kozareva}, {and} \bibinfo{person}{Yue Zhang}} (Eds.). \bibinfo{publisher}{Association for Computational Linguistics}, \bibinfo{address}{Abu Dhabi, United Arab Emirates}, \bibinfo{pages}{8273--8288}.
\newblock
\urldef\tempurl%
\url{https://doi.org/10.18653/v1/2022.emnlp-main.566}
\showDOI{\tempurl}


\bibitem[Tan et~al\mbox{.}(2024)]%
        {tan2024lloco}
\bibfield{author}{\bibinfo{person}{Sijun Tan}, \bibinfo{person}{Xiuyu Li}, \bibinfo{person}{Shishir Patil}, \bibinfo{person}{Ziyang Wu}, \bibinfo{person}{Tianjun Zhang}, \bibinfo{person}{Kurt Keutzer}, \bibinfo{person}{Joseph~E Gonzalez}, {and} \bibinfo{person}{Raluca~Ada Popa}.} \bibinfo{year}{2024}\natexlab{}.
\newblock \showarticletitle{LLoCO: Learning Long Contexts Offline}.
\newblock \bibinfo{journal}{\emph{arXiv preprint arXiv:2404.07979}} (\bibinfo{year}{2024}).
\newblock


\bibitem[Touvron et~al\mbox{.}(2023)]%
        {touvron2023llama}
\bibfield{author}{\bibinfo{person}{Hugo Touvron}, \bibinfo{person}{Louis Martin}, \bibinfo{person}{Kevin Stone}, \bibinfo{person}{Peter Albert}, \bibinfo{person}{Amjad Almahairi}, \bibinfo{person}{Yasmine Babaei}, \bibinfo{person}{Nikolay Bashlykov}, \bibinfo{person}{Soumya Batra}, \bibinfo{person}{Prajjwal Bhargava}, \bibinfo{person}{Shruti Bhosale}, \bibinfo{person}{Dan Bikel}, \bibinfo{person}{Lukas Blecher}, \bibinfo{person}{Cristian~Canton Ferrer}, \bibinfo{person}{Moya Chen}, \bibinfo{person}{Guillem Cucurull}, \bibinfo{person}{David Esiobu}, \bibinfo{person}{Jude Fernandes}, \bibinfo{person}{Jeremy Fu}, \bibinfo{person}{Wenyin Fu}, \bibinfo{person}{Brian Fuller}, \bibinfo{person}{Cynthia Gao}, \bibinfo{person}{Vedanuj Goswami}, \bibinfo{person}{Naman Goyal}, \bibinfo{person}{Anthony Hartshorn}, \bibinfo{person}{Saghar Hosseini}, \bibinfo{person}{Rui Hou}, \bibinfo{person}{Hakan Inan}, \bibinfo{person}{Marcin Kardas}, \bibinfo{person}{Viktor Kerkez}, \bibinfo{person}{Madian Khabsa},
  \bibinfo{person}{Isabel Kloumann}, \bibinfo{person}{Artem Korenev}, \bibinfo{person}{Punit~Singh Koura}, \bibinfo{person}{Marie-Anne Lachaux}, \bibinfo{person}{Thibaut Lavril}, \bibinfo{person}{Jenya Lee}, \bibinfo{person}{Diana Liskovich}, \bibinfo{person}{Yinghai Lu}, \bibinfo{person}{Yuning Mao}, \bibinfo{person}{Xavier Martinet}, \bibinfo{person}{Todor Mihaylov}, \bibinfo{person}{Pushkar Mishra}, \bibinfo{person}{Igor Molybog}, \bibinfo{person}{Yixin Nie}, \bibinfo{person}{Andrew Poulton}, \bibinfo{person}{Jeremy Reizenstein}, \bibinfo{person}{Rashi Rungta}, \bibinfo{person}{Kalyan Saladi}, \bibinfo{person}{Alan Schelten}, \bibinfo{person}{Ruan Silva}, \bibinfo{person}{Eric~Michael Smith}, \bibinfo{person}{Ranjan Subramanian}, \bibinfo{person}{Xiaoqing~Ellen Tan}, \bibinfo{person}{Binh Tang}, \bibinfo{person}{Ross Taylor}, \bibinfo{person}{Adina Williams}, \bibinfo{person}{Jian~Xiang Kuan}, \bibinfo{person}{Puxin Xu}, \bibinfo{person}{Zheng Yan}, \bibinfo{person}{Iliyan Zarov}, \bibinfo{person}{Yuchen
  Zhang}, \bibinfo{person}{Angela Fan}, \bibinfo{person}{Melanie Kambadur}, \bibinfo{person}{Sharan Narang}, \bibinfo{person}{Aurelien Rodriguez}, \bibinfo{person}{Robert Stojnic}, \bibinfo{person}{Sergey Edunov}, {and} \bibinfo{person}{Thomas Scialom}.} \bibinfo{year}{2023}\natexlab{}.
\newblock \bibinfo{title}{Llama 2: Open Foundation and Fine-Tuned Chat Models}.
\newblock
\newblock
\showeprint[arxiv]{2307.09288}~[cs.CL]


\bibitem[Xu et~al\mbox{.}(2023)]%
        {xu2023recomp}
\bibfield{author}{\bibinfo{person}{Fangyuan Xu}, \bibinfo{person}{Weijia Shi}, {and} \bibinfo{person}{Eunsol Choi}.} \bibinfo{year}{2023}\natexlab{}.
\newblock \bibinfo{title}{RECOMP: Improving Retrieval-Augmented LMs with Compression and Selective Augmentation}.
\newblock
\newblock
\showeprint[arxiv]{2310.04408}~[cs.CL]


\bibitem[Yang et~al\mbox{.}(2015)]%
        {yang-etal-2015-wikiqa}
\bibfield{author}{\bibinfo{person}{Yi Yang}, \bibinfo{person}{Wen-tau Yih}, {and} \bibinfo{person}{Christopher Meek}.} \bibinfo{year}{2015}\natexlab{}.
\newblock \showarticletitle{{W}iki{QA}: A Challenge Dataset for Open-Domain Question Answering}. In \bibinfo{booktitle}{\emph{Proceedings of the 2015 Conference on Empirical Methods in Natural Language Processing}}, \bibfield{editor}{\bibinfo{person}{Llu{\'\i}s M{\`a}rquez}, \bibinfo{person}{Chris Callison-Burch}, {and} \bibinfo{person}{Jian Su}} (Eds.). \bibinfo{publisher}{Association for Computational Linguistics}, \bibinfo{address}{Lisbon, Portugal}, \bibinfo{pages}{2013--2018}.
\newblock
\urldef\tempurl%
\url{https://doi.org/10.18653/v1/D15-1237}
\showDOI{\tempurl}


\bibitem[Yang et~al\mbox{.}(2018)]%
        {yang-etal-2018-hotpotqa}
\bibfield{author}{\bibinfo{person}{Zhilin Yang}, \bibinfo{person}{Peng Qi}, \bibinfo{person}{Saizheng Zhang}, \bibinfo{person}{Yoshua Bengio}, \bibinfo{person}{William Cohen}, \bibinfo{person}{Ruslan Salakhutdinov}, {and} \bibinfo{person}{Christopher~D. Manning}.} \bibinfo{year}{2018}\natexlab{}.
\newblock \showarticletitle{{H}otpot{QA}: A Dataset for Diverse, Explainable Multi-hop Question Answering}. In \bibinfo{booktitle}{\emph{Proceedings of the 2018 Conference on Empirical Methods in Natural Language Processing}}, \bibfield{editor}{\bibinfo{person}{Ellen Riloff}, \bibinfo{person}{David Chiang}, \bibinfo{person}{Julia Hockenmaier}, {and} \bibinfo{person}{Jun{'}ichi Tsujii}} (Eds.). \bibinfo{publisher}{Association for Computational Linguistics}, \bibinfo{address}{Brussels, Belgium}, \bibinfo{pages}{2369--2380}.
\newblock
\urldef\tempurl%
\url{https://doi.org/10.18653/v1/D18-1259}
\showDOI{\tempurl}


\bibitem[Zhu et~al\mbox{.}(2024)]%
        {zhu2024accelerating}
\bibfield{author}{\bibinfo{person}{Yun Zhu}, \bibinfo{person}{Jia-Chen Gu}, \bibinfo{person}{Caitlin Sikora}, \bibinfo{person}{Ho Ko}, \bibinfo{person}{Yinxiao Liu}, \bibinfo{person}{Chu-Cheng Lin}, \bibinfo{person}{Lei Shu}, \bibinfo{person}{Liangchen Luo}, \bibinfo{person}{Lei Meng}, \bibinfo{person}{Bang Liu}, {et~al\mbox{.}}} \bibinfo{year}{2024}\natexlab{}.
\newblock \showarticletitle{Accelerating Inference of Retrieval-Augmented Generation via Sparse Context Selection}.
\newblock \bibinfo{journal}{\emph{arXiv preprint arXiv:2405.16178}} (\bibinfo{year}{2024}).
\newblock


\end{thebibliography}
\appendix
\clearpage
\onecolumn
\section{Appendix}
\begin{table*}[!htb]
    \centering
    
    \caption{Results in Match (M) comparing \acrs(-light) to other context compression works. All methods use 5 context passages unless indicated otherwise. $^{\bigstar}$ Method limited to single context. $^{\bigtriangleup}$ upper baseline. $^{\bigtriangledown}$ lower baseline. $^*$ indicates statistical non-significance (p>0.05) with respect to COCOM $\xi$=4.}
    \setlength\tabcolsep{0pt} 
    \begin{tabular*}{\linewidth}{@{\extracolsep{\fill}} cp{0pt}llr *{6}{c}}     
    \toprule
   \multicolumn{2}{c}{\bfseries Decoder} &\bfseries Method & \bf Compression rate ($\xi$) & \multicolumn{6}{c}{\bf Dataset}\\\cmidrule(lr){5-11}
   
   &&  &&  \bfseries NQ & \bfseries TriviaQA & \bfseries HotpotQA & \bfseries ASQA & \bfseries PopQA & \bfseries Average\\
   
    \midrule
   \multirow{5}{*}{\rotatebox{90}{Zero-shot}} & &  AutoCompressor \cite{chevalier2023adapting} $^{\bigstar}$& $\times$ 4 & 0.351 &0.703 & 0.314 &0.574 &0.237  & 0.435\\
  & & ICAE \cite{ge2024incontext} & $\times$ 4 &0.421 &0.784  & 0.293& 0.469& 0.426& 0.479 \\ 
   & & xRAG \cite{cheng2024xrag}$^{\bigstar}$ & & &  & &  & &  \\
   & & \multicolumn{1}{r}{Mistral-7B-v0.2}  & $\times$ 128  &  0.316 &  0.766 &  0.267 & 0.339 & 0.326 & 0.403 \\ 
   & & \multicolumn{1}{r}{Mixtral-8x7b}  & $\times$ 128   & 0.405 &  0.852 & 0.326 & 0.457 & 0.412 & 0.490 \\
    \midrule

   \multirow{9}{*}{\rotatebox{90}{Fine-tuned}}&\multirow{9}{*}{\rotatebox{90}{\textit{Mistral-7B-v0.2}}}  & RAG$^{\bigtriangleup}$ (no compression) & - & 0.637 & 0.917 & 0.544 & 0.665* & 0.543 & 0.661 \\
    & & LLM$^{\bigtriangledown}$ (no context) & -  &0.403 &0.753 & 0.283 & 0.573& 0.208 & 0.444 \\ \cmidrule{3-11} 

    & &  \acrs-light \textit{(ours)} & $\times$ 4 & 0.579 & 0.882 & 0.439 & 0.633* & 0.473 & 0.601 \\
    & & &$\times$ 16 & 0.529 & 0.857 & 0.395 & 0.604 & 0.395 & 0.556 \\
    && &$\times$ 128 &  0.479 & 0.828 & 0.347 & 0.586 & 0.326 & 0.513 \\ \cmidrule{3-11} 
   &  &  \acrs \textit{(ours)} & $\times$ 4& 0.589 & 0.894 & 0.461 & 0.640 &  0.487 & 0.614 \\
   & &  &  $\times$ 16 & 0.577* & 0.886* & 0.456* & 0.633* & 0.478 & 0.606 \\
  &  & &  $\times$ 128 & 0.546 & 0.866 & 0.403 & 0.617* & 0.402 & 0.567 \\
    \bottomrule
    \end{tabular*}
    \label{tab:main_results_match}
\end{table*}

\begin{table}[!h]
    \centering
    \caption{Hyperparameters for Pretraining}
    \setlength\tabcolsep{0pt} 
    \begin{tabular*}{.4\linewidth}{@{\extracolsep{\fill}} l *{1}{c}}    
    \toprule
    \bf Hyperparameter & \bf Assignment  \\
     \midrule
       learning Rate & 1e-4 \\
       lr scheduler type & linear \\
       warmup ratio &  0.05 \\
       weight dacay & 0.1 \\
       overall batch size & 256 \\
       optimizer & AdamW \\
       epochs & 1\\
       LoRa layers & all linear layers\\
       LoRa alpha & 32 \\
       LoRa dropout & 0.1\\
       LoRa $r$ & 16\\
       LoRa bias & None \\
       GPU & 8 x A100 80GB \\
       context max length & 128 \\
       \bottomrule
    \end{tabular*}

    \label{tab:pt_params}
\end{table}

\begin{table}[!h]
    \centering
        \caption{Hyperparameters for Fine-tuning}
    \setlength\tabcolsep{0pt} 
    \begin{tabular*}{.4\linewidth}{@{\extracolsep{\fill}} l *{1}{c}} 
    \toprule
    \bf Hyperparameter & \bf Assignment  \\
     \midrule
       learning Rate & 1e-4 \\
       lr scheduler type & linear \\
       warmup ratio &  0.05 \\
       weight dacay & 0.1 \\
       overall batch size & 64 \\
       optimizer & AdamW \\
       epochs & 2\\
       LoRa layers & all linear layers\\
       LoRa alpha & 32 \\
       LoRa dropout & 0.1\\
       LoRa $r$ & 16\\
       LoRa bias & None \\
       GPU & 8 x A100 80GB \\
       retriever(s)  &  SPLADE-v3 (+ DeBERTa-v3) \\
       num passages & 5 \\
       \bottomrule
    \end{tabular*}
    \label{tab:ft_params}
\end{table}

\begin{table}[!h]
    \centering
        \caption{Datasets contained in the multi-dataset collection used for fine-tuning our \acrs(-light). We filtered out queries with more than 128 tokens and labels of more than 64 tokens.}
    \setlength\tabcolsep{0pt} 
    \begin{tabular*}{.4\linewidth}{@{\extracolsep{\fill}} l *{1}{r}} 
    \toprule
    \bf Dataset & \bf Number examples  \\
       \midrule
       NQ & 87,925\\
       MSMARCO                & 100,000 \\
        Adversarial QA         & 30,000  \\
        HotpotQA               & 88,869  \\
        WikiQA                 & 873     \\
        SciQ                   & 11,679  \\
        ASQA                   & 4,353   \\
        Wiki QA  & 61,817  \\
        Freebase               & 20,358  \\
        SQuAD                  & 87,599  \\
        \midrule
       Total & 493,473 \\
       \bottomrule
    \end{tabular*}
    \label{tab:multi_dataset}
\end{table}

\appendix

\end{document}